\documentclass{tlp}
\usepackage{aopmath}

\usepackage{latexsym} 
\usepackage{amssymb}

\usepackage{xspace}
\usepackage{subfigure}

\usepackage{url}

\usepackage{epsfig}
\usepackage{graphicx}

\usepackage{times}

\newcommand{\clasp}{{\sc clasp}\xspace}
\newcommand{\claspD}{{\sc claspD}\xspace}
\newcommand{\claspf}{{\sc claspfolio}\xspace}
\newcommand{\claspfolio}{\claspf}
\newcommand{\cmodels}{{\sc cmodels}\xspace}
\newcommand{\dlv}{{\sc DLV}\xspace}
\newcommand{\idp}{{\sc idp}\xspace}

\newcommand{\smodels}{\sc{smodels}\xspace}
\newcommand{\dors}{\sc{dors}\xspace} 

\newcommand{\gringo}{{\sc GrinGo}\xspace}
\newcommand{\measp}{{\sc me-asp}\xspace}
\newcommand{\meclasp}{{\sc me-clasp}\xspace}

\newcommand{\apc}{{\sc apc}\xspace}
\newcommand{\furia}{{\sc furia}\xspace}
\newcommand{\jqo}{{\sc j48}\xspace}
\newcommand{\mlr}{{\sc mlr}\xspace}
\newcommand{\nn}{{\sc nn}\xspace}
\newcommand{\svm}{{\sc svm}\xspace}

\newcommand{\sota}{\textsc{sota}\xspace}

\newcommand{\NP}{{\em NP}\xspace}
\newcommand{\BNP}{{\em Beyond NP}\xspace}

\newcommand{\GR}{\ensuremath{Ground(r)}\xspace}
\newcommand{\R}{\ensuremath{r}}

\newcommand{\Or}{\ensuremath{\vee}}
\newcommand{\derives}{\ensuremath{\mbox{\,$:$--}\,}\xspace}

\newcommand{\naf}{\ensuremath{\mathtt{not\ }}}

\newcommand{\p}{\ensuremath{{\cal P}}\xspace}
\newcommand{\GP}{\ensuremath{Ground(\p)}\xspace}
\newcommand{\BP}{\ensuremath{B_{\p}}\xspace}
\newcommand{\UP}{\ensuremath{U_{\p}}\xspace}

\newcommand{\TS}[1]{\textsc{ts}$_{#1}$\xspace}

\newcommand{\TSU}{\textsc{ts}\xspace}
\newcommand{\MOD}[1]{{\textsc{mod$_{#1}$}}\xspace}

\newcommand{\measpC}[1]{{\measp(#1)}\xspace}

\newcommand{\citeASP}{\cite{gelf-lifs-88,eite-etal-97f,mare-trus-98sm,niem-98,lifs-99a,gelf-lifs-91,bara-2002}\xspace}

\begin{document}
\bibliographystyle{acmtrans}


\hyphenation{La-by-ri-nth}

\title[A Multi-Engine Approach to Answer Set Programming]{A Multi-Engine Approach to Answer Set Programming\footnote{This is an extended and revised version of~\cite{mara-etal-2012-iclp,mara-etal-2012-jelia}.}}

\author {Marco Maratea \and Luca Pulina \and Francesco Ricca}

\author[M. Maratea et. al]
{Marco Maratea$^1$, Luca Pulina$^{2}$, Francesco Ricca$^3$\\
$^1$DIBRIS, Universit{\`a} degli Studi di Genova, Viale F.Causa 15, 16145 Genova, Italy\\
$^2$POLCOMING, Universit{\`a} degli Studi di Sassari,
Viale Mancini 5, 07100 Sassari, Italy\\
$^3$Dipartimento di Matematica, Universit{\`a} della Calabria, Via P. Bucci, 87030 Rende, Italy\\
\email{marco@dist.unige.it, lpulina@uniss.it, ricca@mat.unical.it}
}



\submitted{12 July 2012}
\revised{18 December 2012}
\accepted{4 June 2013}

\maketitle

\label{firstpage}

\begin{abstract}
Answer Set Programming (ASP) is a truly-declarative programming paradigm proposed in the area of non-monotonic reasoning and logic programming, that
has been recently employed in many applications. The development of efficient ASP systems is, thus, crucial.
Having in mind the task of improving the solving methods for ASP, 
there are two usual ways to reach this goal: $(i)$
extending state-of-the-art techniques and ASP solvers, or $(ii)$
designing  a new ASP solver from scratch. An alternative to these
trends is to build on top of state-of-the-art solvers, and to apply
machine learning techniques for choosing automatically the ``best''
available solver on a per-instance basis.

In this paper we pursue this latter direction. We first define a set
of cheap-to-compute syntactic features that characterize several
aspects of ASP programs. Then, we apply classification methods that,
given the features of the instances in a {\sl training} set and the
solvers' performance on these instances, inductively learn algorithm
selection strategies to be applied to a {\sl test} set.  We report the
results of a number of experiments considering solvers and different
training and test sets of instances taken from the ones submitted to
 the ``System Track'' of the 3rd ASP Competition. Our analysis shows
that, by applying machine learning techniques to ASP solving, it is
possible to obtain very robust performance: our approach can solve 
more instances compared with any solver that entered the 3rd ASP Competition.
(To appear in Theory and Practice of Logic Programming (TPLP).)
\end{abstract}
\begin{keywords}
 Answer Set Programming, Automated Algorithm Selection, Multi-Engine solvers 
\end{keywords}

\section{Introduction}\label{sec:intro}

Answer Set Programming~\citeASP (ASP) is a truly-declarative programming paradigm proposed in the area of non-monotonic reasoning and logic programming.
The idea of ASP is to represent a given computational
problem by a logic program whose answer sets correspond to solutions,
and then use a solver to find such solutions \cite{lifs-99a}.
The language of ASP is very expressive, indeed
all problems in the second level of the polynomial hierarchy can be 
expressed in ASP \cite{eite-etal-97f}. Moreover, in the last years ASP has been employed in many applications, see, e.g.,~\cite{noge-etal-2001,bara-2002,broo-etal-2007,frie-08-techrep,gebs-etal-2011-bio,bald-lier-12}, and even in industry~\cite{ricc-etal-2008-jlc,ielp-etal-09-idum-lpnmr,rull-etal-2009,ricc-etal-2010-tplp-gioia,marczak-etal-sigmod10,smara-etal-popl11}.
The development of efficient ASP systems 
is, thus, a crucial task, made even more challenging by existing and new-coming applications.

Having in mind the task of improving the robustness, i.e., the ability
to perform well across a wide set of problem domains, and the
efficiency, i.e., the quality of solving a high number of instances,
of solving methods for Answer Set Programming (ASP), 
it is possible to extend existing state-of-the-art techniques implemented in ASP
solvers, or design from scratch a new ASP system with powerful
techniques and heuristics.
An alternative to these trends is to build on top of
state-of-the-art solvers, leveraging on a number of efficient 
ASP systems, e.g.,~\cite{simo-etal-2002,leon-etal-2002-dlv,giun-etal-2006,gebs-etal-2007-ijcai,mari-etal-2008,janh-etal-2009}, %
and applying machine learning techniques for inductively choosing,
among a set of available ones, the ``best'' solver on the basis of the
characteristics, called {\sl features}, of the input program. This
approach falls in the framework of the \textsl{algorithm selection
  problem}~\cite{rice-76}.  Related approaches, following this
per-instance selection, have been exploited for solving propositional
satisfiability (SAT), e.g.,~\cite{xu-etal-2008}, and Quantified SAT
(QSAT), e.g.,~\cite{puli-tacc-07} problems.  In ASP, an approach for
selecting the ``best'' {\clasp} internal configuration is followed
in~\cite{gebs-etal-2011-claspfolio}, while another approach that
imposes learned heuristics ordering to {\smodels} is~\cite{bald-11}.

In this paper we pursue this direction, and propose a
multi-engine approach to ASP solving. We first define a set of
cheap-to-compute syntactic features that describe several
characteristics of ASP programs, paying particular attention to ASP
peculiarities. We then compute such features for the grounded version
of all benchmarks submitted to the ``System Track'' of the 3rd ASP
Competition~\cite{cali-etal-2011-syscomp-tplp}
falling in the ``{\NP}'' and ``{\BNP}'' categories of the competition:
this track is well suited for our study given that $(i)$ contains many
ASP instances, $(ii)$ the language specification, ASP-Core, is a
common ASP fragment such that $(iii)$ many ASP systems can deal with
it.

   Then, we apply classification methods that, starting from the
   features of the instances in a {\sl training} set, and the solvers'
   performance on these instances, inductively learn general
   algorithm selection strategies to be applied to a {\sl test}
   set. We consider six well-known multinomial classification
   methods, some of them considered in~\cite{puli-tacc-07}. We
   perform a number of analysis considering different training and
   test sets.
   Our experiments show that it is
   possible to obtain a very robust performance, by solving many
   more instances than all the solvers that
   entered the 3rd ASP Competition and {\dlv}~\cite{leon-etal-2002-dlv}.

The paper is structured as follow. Section~\ref{sec:prelim} contains
preliminaries about ASP and classification methods. Section~\ref{sec:benchs}
then describes our benchmark setting, in terms of dataset and solvers
employed. Section~\ref{sec:main} defines how features and solvers have
been selected, and presents the classification methods
employed. Section~\ref{sec:exp} is dedicated to the performance analysis, while
Section~\ref{sec:related} and~\ref{sec:conclusion} end the paper with
discussion about related work and conclusions, respectively.

\section{Preliminaries}\label{sec:prelim}
In this section we recall some preliminary notions concerning
Answer Set Programming and 
machine learning techniques for algorithm selection.

\subsection{Answer Set Programming}\label{sec:asp}

In the following, we recall both the syntax and semantics of ASP.
The presented constructs are included in ASP-Core~\cite{cali-etal-2011-syscomp-tplp}, 
which is the language specification that was originally introduced in 
the 3rd ASP Competition~\cite{cali-etal-2011-syscomp-tplp}
as well as the one employed in our experiments (see Section~\ref{sec:benchs}).
Hereafter, we assume the reader is familiar with logic programming conventions, 
and refer the reader to~\cite{gelf-lifs-91,bara-2002,gelf-leon-02} 
for complementary introductory material on ASP, and 
to \cite{asp-comp-11-web} 
for obtaining the full specification of ASP-Core.

\paragraph{\bf Syntax.}
A variable or a constant is a {\em term}.  An {\em atom} is
$p(t_{1},..., t_{n})$, where $p$ is a {\em predicate} of arity $n$
and $t_{1},..., t_{n}$ are terms.  A {\em literal} is either a 
{\em positive~literal} $p$ or a {\em negative~literal} $\naf p$, 
where $p$ is an atom.
A {\em (disjunctive) rule} $r$ is of the form:
\[
a_1\ \Or\ \cdots\ \Or\ a_n\ \derives\ 
        b_1,\cdots, b_k,\ 
        \naf\ b_{k+1},\cdots,\ \naf\ b_m.
\]
\noindent where $a_1,\ldots ,a_n,b_1,\ldots ,b_m$ are atoms.
The disjunction $a_1\lor\ldots\lor a_n$ is the {\em head} of $r$, while
the conjunction $b_1 , \ldots,  b_k, \naf b_{k+1} , \ldots, \naf b_m$ 
is the {\em body} of $r$.
We denote by $H(r)$ the set of atoms occurring in the head of $r$,
and we denote by $B(r)$ the set of body literals. 
A rule s.t. $|H(r)| = 1$ (i.e., $n=1$) is called a {\em normal rule}; 
if the body is empty (i.e.,\ $k=m=0$) it is called a {\em fact}
(and the $\derives$ sign is omitted); if $|H(r)| = 0$ (i.e., $n=0$)
is called a {\em constraint}.
A rule $r$ is {\em safe} if each variable appearing in $r$ appears 
also in some positive body literal of $r$.

An {\em ASP program} \p is a finite set of safe rules. 
A $\naf\!$-free (resp., $\lor$-free) program 
is called {\em positive} (resp., {\em normal}).  
A term, an atom, a literal, a rule, or a
program is {\em ground} if no variable appears in it.

\paragraph{\bf  Semantics.}
Given a program \p, the \emph{Herbrand Universe} \UP
is the set of all constants appearing in \p, and 
the \emph{Herbrand Base} \BP is the set of all possible ground atoms 
which can be constructed from the predicates appearing in \p 
with the constants of \UP.
Given a rule $r$, \GR denotes the set of rules obtained 
by applying all possible substitutions from the variables in $r$ to elements of \UP. 
Similarly, given a program \p, the {\em ground instantiation} of \p 
is $\GP = \bigcup_{\R \in \p} \GR$.

An {\em interpretation} for a program \p is a subset $I$ of \BP. 
A ground positive literal $A$ is true (resp., false) w.r.t. $I$ if $A \in I$ (resp.,  $A \not\in I$). 
A ground negative literal $\naf A$ is true w.r.t. $I$ if $A$ is false w.r.t. $I$; 
otherwise $\naf A$ is false w.r.t. $I$.

The answer sets of a program \p are defined in two steps
using its ground instantiation:
first the answer sets of positive disjunctive programs are defined;
then the answer sets of general programs are defined by a reduction to
positive ones and a stability condition.

Let  $r$ be a ground rule, the head of $r$ is true w.r.t. $I$ if $H(r) \cap I \neq \emptyset$. 
The body of $r$ is true w.r.t. $I$ if all body literals of  $r$ are true w.r.t. $I$, 
otherwise the body of $r$ is false w.r.t. $I$.
The rule  $r$ is {\em satisfied} (or true) w.r.t. $I$ 
if its head is true w.r.t.  $I$ or its body is false w.r.t. $I$.

Given a {\em ground positive} program $P_g$, an {\em answer set}
for $P_g$ is a subset-minimal interpretation $A$ for $P_g$ such that every rule $r \in P_g$ is true w.r.t. $A$
(i.e., there is no other interpretation $I \subset A$ that satisfies all the rules of $P_g$).

Given a {\em ground} program  $P_g$ and an interpretation $I$, the (Gelfond-Lifschitz) 
{\em reduct}~\cite{gelf-lifs-91} of $P_g$ w.r.t. $I$ is the positive program $P_g^I$, 
obtained from $P_g$ by 
$(i)$ deleting all rules $r \in P_g$ whose negative body is false w.r.t.\ $I$, and 
$(ii)$ deleting the negative body from the remaining rules of $P_g$.

An answer set (or stable model) of a general program \p is an interpretation $I$ of \p \ such
that $I$ is an answer set of $\GP^I$. \\

As an example consider the program $\p = \{$
  $a \Or b \derives c$., 
  $b \derives \naf a, \naf c$.,
  $a \Or c \derives \naf b$.,
  $k \derives a$.,
  $k \derives b$. $\}$ 
and $I = \{ b, k \}$. The reduct $\p^I$ is $\{a \Or b \derives c$., $b$. $k \derives a$., $k \derives b$.$\}$. 
$I$ is an answer set of $\p^I$, and for this reason
it is also an answer set of \p.

\subsection{Multinomial Classification for Algorithm Selection}
\label{sec:multiasp}

With regard to empirically hard problems, there is rarely a best
algorithm to solve a given combinatorial problem, while it is often
the case that different algorithms perform well on different problem
instances. In this work
we rely on a per-instance selection algorithm in which, given a set of
{\it features} --i.e., numeric values that represent particular
characteristics of a given instance-- it is possible to choose the
best (or a good) algorithm among a pool of them --in our case, 
ASP solvers. In order to make such a selection in an
automatic way, we model the problem using {\it multinomial
  classification} algorithms, i.e., machine learning techniques that
allow automatic classification of a set of instances, given some
instance features.

In more detail, in multinomial classification we are given a set of
patterns, i.e., input vectors $X = \{\underline{x}_1, \dots
\underline{x}_k\}$ with $\underline{x}_i \in \mathbb{R}^{n}$, and a
corresponding set of labels, i.e., output values $Y \in \{1, \dots,
m\}$, where $Y$ is composed of values representing the $m$ classes
of the multinomial classification problem. In our modeling, the $m$
classes are $m$ ASP solvers. We think to the labels as generated by
some unknown function $f : \mathbb{R}^{n}\rightarrow \{1, \dots, m\}$
applied to the patterns, i.e., $f(\underline{x}_i) = y_i$ for $i \in
\{1, \dots , k\}$ and $y_i \in \{1, \dots, m\}$.  Given a set of
patterns $X$ and a corresponding set of labels $Y$, the task of a
multinomial classifier $c$ is to extrapolate $f$ given $X$ and $Y$,
i.e., construct $c$ from $X$ and $Y$ so that when we are given some
$\underline{x}^{\star} \in X$ we should ensure that
$c(\underline{x}^{\star})$ is equal to
$f(\underline{x}^{\star})$. This task is called {\it training}, and
the pair $(X,Y)$ is called the {\it training set}.  

\begin{table}[t!]
\setlength{\tabcolsep}{\tabcolsep}
\begin{tabular*}{0.58\columnwidth}{l|l|cc}

\hline
{\bf Problem} & {\bf Class} & {\bf \#Instances}   \\
\hline \hline
DisjunctiveScheduling & {\NP} & 10 \\
GraphColouring & {\NP} & 60 \\
HanoiTower & {\NP} & 59 \\
KnightTour & {\NP} & 10 \\
MazeGeneration & {\NP} & 50 \\
Labyrinth & {\NP} & 261 \\
MultiContextSystemQuerying & {\NP} & 73 \\
Numberlink & {\NP} & 150 \\
PackingProblem & {\NP} & 50 \\
SokobanDecision & {\NP} & 50 \\
Solitaire & {\NP} & 25 \\
WeightAssignmentTree & {\NP} & 62 \\
\hline
MinimalDiagnosis & {\BNP} & 551\\
StrategicCompanies & {\BNP} & 51 \\
\hline
Total & \ & 1462\\
\hline
\end{tabular*}\caption{\small Problems and instances.}\label{tab:benchmarks}
\end{table}

\section{Benchmark Data and Settings}
\label{sec:benchs}

In this section we report 
the benchmark settings employed in this work,
which is needed for properly introducing the techniques
described in the remainder of the paper.
In particular, we report some data concerning:
benchmark problems, instances and ASP solvers employed,
as well as the hardware platform, and the execution settings
for reproducibility of experiments.

\subsection{Dataset}

The benchmarks considered for the experiments
belong to the suite of the 3rd ASP Competition~\cite{cali-etal-2011-syscomp}.  
This is a large and heterogeneous suite of hard benchmarks encoded in ASP-Core,
which was already employed for evaluating the performance of state-of-the-art ASP solvers.
That suite includes planning domains, temporal and
spatial scheduling problems, combinatorial puzzles, graph problems, and
a number of application domains, i.e., databases, information extraction and molecular biology field.%
\footnote{An exhaustive description of the benchmark problems can be found in~\cite{asp-comp-11-web}.}
In more detail, we have employed the encodings used in the System Track of the competition,
and all the problem instances made {\em available} (in form of facts) 
from the contributors of the problem submission stage of the competition,
which are available from the competition website~\cite{asp-comp-11-web}. 
Note that this is a superset of the instances actually selected for running (and, thus {\em evaluated} in) the competition itself.  
Hereafter, with {\em instance} we refer to the complete input program (i.e., encoding+facts) 
to be fed to a solver for each instance of the problem to be solved.

The techniques presented in this paper are conceived for dealing with propositional programs, 
thus we have grounded all the mentioned instances 
by using \gringo (v.3.0.3)~\cite{gebs-etal-2007-gringo}
to obtain a setup very close to the one of the competition. %
We considered only computationally-hard benchmarks,
corresponding to all problems belonging to the categories {\NP} and {\BNP} of the competition.
The dataset is summarized in Table~\ref{tab:benchmarks}, which
also reports the complexity classification 
and the number of available instances for each problem.

\subsection{Executables and Hardware Settings}
We have run  all the ASP solvers that 
entered the System Track of the 3rd ASP Competition~\cite{cali-etal-2011-syscomp}
with the addition of \dlv~\cite{leon-etal-2002-dlv} 
(which did not participate in the competition since it is developed by the organizers of the event).
In this way we have covered --to the best of our knowledge--
all the state-of-the-art solutions fitting the benchmark settings.
In detail, we have run:
\clasp~\cite{gebs-etal-2007-ijcai}, \claspD~\cite{dres-etal-2008-KR}, \claspfolio~\cite{gebs-etal-2011-claspfolio},
\idp~\cite{witt-etal-2008-idp}, \cmodels~\cite{lier-2005-lpnmr},  {\sc sup}~\cite{lier-2008-sup}, 
{\sc Smodels}~\cite{simo-etal-2002},
and several solvers from both the {\sc lp2sat}~\cite{janh-2006-journal-lp2sat} and  {\sc lp2diff}~\cite{janh-etal-2009} 
families, namely: {\sc lp2gminisat}, {\sc lp2lminisat}, {\sc lp2lgminisat}, {\sc lp2minisat}, 
{\sc lp2diffgz3}, {\sc lp2difflgz3}, {\sc lp2difflz3}, and {\sc lp2diffz3}.
More in detail, \clasp\ is a native ASP solver relying on conflict-driven nogood learning;
\claspD\ is an extension of \clasp\ that is able to deal with disjunctive logic programs,
while \claspfolio\ exploits machine learning techniques in order to choose
the best-suited execution options of \clasp; 
\idp is a finite model generator for extended first-order logic theories, 
which is based on {\em MiniSatID}~\cite{mari-etal-2008};
{\sc Smodels} is one of the first robust native ASP solvers that have been made available to the community;
\dlv~\cite{leon-etal-2002-dlv} is one of the first systems able to cope with disjunctive programs;
\cmodels\ exploits a SAT solver as a search engine for enumerating models, 
and also verifies model minimality with SAT, whenever needed; {\sc sup}\ exploits nonclausal constraints, and
can be seen as a combination of the computational ideas behind \cmodels\ and {\sc Smodels}; 
the {\sc lp2sat} family employs several variants (indicated by the trailing {\sc g}, {\sc l} and {\sc lg}) 
of a translation strategy to SAT and resorts to {\sc MiniSat}~\cite{een-sore-03}
for actually computing the answer sets; the {\sc lp2diff} family translates programs 
in difference logic over integers~\cite{smt-lib-web} 
and exploit {\em Z3}~\cite{mend-bjor-2008-z3} as underlying solver
(again, {\sc g}, {\sc l} and {\sc lg} indicate different translation strategies).
{\dlv} was run with default settings, while remaining solvers were run on the same configuration (i.e., parameter settings) as in the competition.

Concerning the hardware employed and the execution settings, 
all the experiments were carried out on CyberSAR~\cite{cybersar}, 
a cluster comprised of 50 Intel Xeon E5420 blades 
equipped with 64 bit GNU Scientific Linux 5.5. 
Unless otherwise specified, the resources granted to 
the solvers are 600s of CPU time and 2GB of memory. 
Time measurements were carried out using the
\texttt{time} command shipped with GNU Scientific Linux 5.5.

\section{Designing a Multi-Engine ASP Solver}\label{sec:main}

The design of a multi-engine solver 
involves several steps: $(i)$ design of (syntactic) features that are both significant 
for classifying the instances and cheap-to-compute (so that the classifier can be fast and accurate); 
$(ii)$ selection of solvers that are representative of the state of the art 
(to be able to possibly obtain the best performance in any considered instance); and
$(iii)$ selection of the classification algorithm, and 
fair design of training and test sets, to obtain a robust and unbiased classifier.

In the following, we describe the choices we have made for designing
\measp, which is our multi-engine solver for ground ASP programs.

\subsection{Features}
\label{sub:feature}

Our features selection process started by considering a very wide set of candidate features that correspond,
in our view, to several characteristics of an ASP program that, in principle, should be taken into account.

The features that we compute for each ground program are divided into four groups (such a 
categorization is borrowed from~\cite{nude-etal-2004}):
\begin{itemize}
\item {\bf Problem size features}: number of rules $r$, number of atoms $a$, 
ratios $r/a$, $(r/a)^2$, $(r/a)^3$ and ratios reciprocal $a/r$, $(a/r)^2$ and $(a/r)^3$.
This type of features are considered to give an idea of what is the size of the ground program. 
\item {\bf Balance features}: ratio of positive and negative atoms in each body, and ratio of positive and negative
occurrences of each variable; fraction of unary, binary and ternary rules. This type of
features can help to understand what is the ``structure'' of the analyzed program.
\item {\bf ``Proximity to horn" features}: fraction of horn rules and number of atoms occurrences in horn rules. These features can give an indication on ``how much'' a program is close to be horn: this can be helpful, 
since some solvers may take advantage from this setting (e.g., minimum or no impact of completion~\cite{clar-78} when applied).
\item {\bf ASP peculiar features}: number of true
and disjunctive facts, fraction of normal rules and constraints, head sizes, occurrences
of each atom in heads, bodies and rules, occurrences of true negated atoms in heads,
bodies and rules; Strongly Connected Components (SCC) sizes, number of Head-Cycle Free (HCF) and non-HCF components, degree of support for non-HCF components. 
\end{itemize}


For the features implying distributions, e.g., ratio of positive and
negative atoms in each body, atoms occurrences in horn rules, and head
sizes, five numbers are considered: minimum, 25\% percentile, median,
75\% percentile and maximum. The five numbers are considered given
that we can not a-priori consider the distributions to be Gaussians, thus mean
and variance are not that informative.

The set of features reported above seems to be adequate for describing 
an ASP program.%
\footnote{Observations concerning existing proposals are reported in Section~\ref{sec:related}.}
On the other hand, we have to consider that the time spent computing the features 
will be integral part of our solving process: the risk is to spend too much time in
calculating the features of a program. This component of the solving process 
could result in a significant overhead in the solving time in case of instances 
that are easily solved by (some of) the engines, or can even cause
a time out on programs otherwise solved by (some of) the engines within the time limit.

Given these considerations, our final choice is to consider syntactic features that are
cheap-to-compute, i.e., computable in linear time in the size of the
input, also given that in previous work (e.g.,~\cite{puli-tacc-07})
syntactic features have been profitably used for characterizing
(inherently) ground instances. 
To this end, we implemented a tool able to compute the above-reported set of features
and conducted some preliminary experiments on all the benchmarks 
we were able to ground with {\gringo} in less than 600s: 
1425 instances out of a total of 1462, of which 823 out of 860 NP instances.%
\footnote{The exceptions are 10 and 27
  instances of DisjunctiveScheduling and PackingProblem,
  respectively.} 
On the one hand, the results confirmed the need for avoiding the computation of
``expensive'' features (e.g., SCCs): indeed, in this setting we could compute 
the whole set of features only for 665 \NP instances within 600s; and, 
on the other hand, the results helped us in selecting a set of ``cheap'' features
that are sufficient for obtaining a robust and efficient multi-engine system.
In particular, the features that we selected are a subset of the ones reported above:

\begin{itemize}
\item {\bf Problem size features}: number of rules $r$, number of atoms $a$, ratios
$r/a$, ${(r/a)}^2$, $(r/a)^3$ and ratios reciprocal $a/r$, ${(a/r)}^2$
and ${(a/r)}^3$;
\item {\bf Balance features}: fraction of unary, binary
and ternary rules;
\item {\bf ``Proximity to horn'' features}: fraction of
horn rules;
\item {\bf ASP peculiar features}: number of true and disjunctive facts,
fraction of normal rules and constraints $c$.
\end{itemize}

This final choice of features, together with some of their combinations (e.g., $c/r$), amounts for a total
of 52 features. 
Our tool for extracting features from ground programs can then
compute all these features (in less than 600s) for 1371 programs out of 1462. 
The distribution of the CPU times for extracting features is
characterized by the following five numbers: 0.24s, 1.74s, 2.40s,
4.37s, 541.92s. It has to be noticed that high CPU times correspond
to extracting features for ground programs whose size is in the order of GigaBytes.
Our set of chosen features is relevant, as will be shown in Section~\ref{sec:exp}.

\subsection{Solvers Selection}
\label{sub:solver}

\begin{table*}[t!]
\small
\begin{center}
\setlength{\tabcolsep}{1.5\tabcolsep}

\begin{tabular*}{0.805\columnwidth}{l|r|r||l|r|r}
\hline
\multicolumn{1}{c|}{\bf Solver} & \multicolumn{1}{|c|}{\bf Solved} & \multicolumn{1}{|c||}{\bf Unique} & \multicolumn{1}{|c|}{\bf Solver} & \multicolumn{1}{|c|}{\bf Solved} & \multicolumn{1}{|c}{\bf Unique} \\ 
\hline
\hline
\textsc{clasp} & 445 & 26 & \textsc{lp2diffz3} & 307 & -- \\ 
\textsc{cmodels} & 333 & 6 & \textsc{lp2sat2gminisat} & 328 & --  \\ 
\textsc{dlv} & 241 & 37 & \textsc{lp2sat2lgminisat} & 322 & -- \\ 
\textsc{idp} & 419 & 15 & \textsc{lp2sat2lminisat} & 324 & -- \\ 
\textsc{lp2diffgz3} & 254 & -- & \textsc{lp2sat2minisat} & 336 & -- \\ 
\textsc{lp2difflgz3} & 242 & -- & \textsc{smodels} & 134 & -- \\ 
\textsc{lp2difflz3} & 248 & -- & \textsc{sup} & 311 & 1 \\ 
\hline
\end{tabular*}
\caption{\small Results of a pool of ASP solvers on the {\NP} instances of
  the 3rd ASP Competition. The table is organized as follows: column
  ``{\bf Solver}'' reports the solver name, column ``{\bf Solved}'' reports the
  total amount of instances solved with a time limit of 600 
  seconds, and, finally, in column ``{\bf Unique}'' we report the total
  amount of uniquely solved instances by the corresponding solver.}
\label{tab:unique}
\end{center}
\end{table*}

The target of our selection is to collect a pool of solvers that is
representative of the state-of-the-art solver ({\sc sota}), i.e.,
considering a problem instance, the oracle that always fares the best
among available solvers. 
Note that, in our settings, the various engines available employ (often substantially) 
different evaluation strategies, and (it is likely that) different engines behaves better in different domains, 
or in other words, the engines' performance is ``orthogonal''. 
As a consequence one can find that there are solvers that solves a significant 
number of instances uniquely (i.e., instances solved by only one solver),
that have a characteristic performance and are a fundamental component of the \sota. 
Thus, a pragmatic and reasonable choice, given that we want to solve as much instances as possible, 
is to consider a solver only if it solves a reasonable amount of instances uniquely, 
since this solver cannot be, in a sense, subsumed performance-wise by another behaving similarly.

In order to select the engines we ran preliminary experiments, and 
we report the results (regarding the \NP class) in Table~\ref{tab:unique}. 
Looking at the table, first we notice that we do not report results related 
to both {\claspD} and \claspfolio. Concerning the results of \claspD, 
we report that --considering the \NP class-- its performance, in terms
of solved instances, is subsumed by the performance of \clasp. 
Considering the performance of {\claspfolio},
we exclude such system from this analysis because we consider it as a
yardstick system, i.e., we will compare its performance against
the performance of {\measp}.

Looking at Table~\ref{tab:unique}, we can see that only 4 solvers out
of 16 are able to solve a noticeable amount of instances {\it
  uniquely}, namely \clasp, \cmodels, \dlv, and \idp.%
\footnote{The picture of uniquely solved instances does not change even considering the entire family of \textsc{lp2sat}
(resp. \textsc{lp2diff}) as a single engine that has the best performance among its variants.}
Concerning {\BNP} instances, we report that only three solvers are able to cope
with such class of problems, namely \claspD, \cmodels, and \dlv. 
Considering that both \cmodels and \dlv are involved in the previous selection, 
that \claspD has a performance that does not overlap with the other two in {\BNP} instances, 
the pool of engines used in {\measp} will be composed of 5 solvers, namely \clasp, \claspD, \cmodels, \dlv, and \idp.

The experiments reported in Section~\ref{sec:exp} confirmed 
that this engine selection policy is effective in practice considering the ASP state of the art.
Nonetheless, it is easy to see that in scenarios where the performance
of most part of the available solvers is very similar on a common pool
of instances, i.e., their performance is not ``orthogonal'', choosing a
solver for the only reason it solves a reasonable amount of instances uniquely
may be not an effective policy. Indeed, the straightforward
application of that policy to ``overlapping'' engines could result in
discarding the best ones, since it is likely that several of them can
solve the same instances.
An effective possible extension of the selection policy
presented above to deal with overlapping engines is to remove {\it
  dominated} solvers, i.e., a solver $s$ dominates a solver $s'$ if
the set of instances solved by $s$ is a superset of the instances
solved by $s'$. Ties are broken choosing the solver that spends the
smaller amount of CPU time. If the resulting pool of engines is still
not reasonably distinguishable, i.e., there are not enough uniquely solved instances by each engine of the pool, then one may compute such pool, say $E$, as follow: starting from the empty set ($E = \emptyset$), 
and trying iteratively to add engine candidates to $E$ from the one that solves 
more instances, and faster, to the less efficient.  At each iteration, an engine $e$ is added
to $E$ if both the set of uniquely solved instances by the engines in
$E\cup\{e\}$ is larger than in $E$, and the resulting set $E\cup\{e\}$
is reasonably distinguishable.

We have applied the above extended policy, that is to be considered as
a pragmatic strategy more than a general solution, obtaining good
results in a specific experiment with overlapping engines; more
details will be found in Section~\ref{sub:discussion}.

\subsection{Classification Algorithms and Training}\label{subsec:classalg}

In the following, we briefly review the classifiers that we use in our
empirical analysis.  Considering the wide range of multinomial
classifiers described in the scientific literature, we test a subset
of algorithms, some of them considered in~\cite{puli-tacc-07}.
Particularly, we can limit our choice to the classifiers able to deal
with numerical attributes (the features) and multinomial class labels
(the engines). Furthermore, in order to make our approach as general
as possible, our desiderata is to choose classifiers that allow us to
avoid ``stringent'' assumptions on the features distributions, e.g.,
hypotheses of normality or independence among the features. At the
end, we also prefer classifiers that do not require complex parameter
tuning, e.g., procedures that are more elaborated than standard
parameters grid search. The selected classifiers are listed in the
following:

\begin{itemize}
\item \textbf{Aggregation Pheromone density based pattern
  Classification} ({\apc}): it is a pattern classification algorithm
  modeled on the ants colony behavior and distributed adaptive
  organization in nature. Each data pattern is considered as an ant,
  and the training patterns (ants) form several groups or colonies
  depending on the number of classes present in the data set. A new
  test pattern (ant) will move along the direction where average
  aggregation pheromone density (at the location of the new ant)
  formed due to each colony of ants is higher and, hence, eventually it
  will join that colony. We refer the reader
  to~\cite{halder2009aggregation} for further details.
\item \textbf{Decision rules} ({\furia}): a classifier providing a set
  of rules that generally takes the form of a Horn clause wherein the
  class labels is implied by a conjunction of some attributes; we use
  \furia~\cite{hhn2009furia} to induce decision rules.
\item \textbf{Decision trees} ({\jqo}): a classifier arranged in a
  tree structure, and used to discover decision rules. Each
  inner node contains a test on some attributes, and each leaf node
  contains a label; we use {\jqo}, an optimized implementation of {\sc
    c4.5}~\cite{quinlan1993c4}.
\item \textbf{Multinomial Logistic Regression} ({\mlr}): a classifier
  providing a hyperplane of the hypersurfaces that separate the class
  labels in the feature space; we use the inducer described
  in~\cite{le1992ridge}.
\item \textbf{Nearest-neighbor} ({\nn}): it is a classifier yielding
  the label of the training instance which is closer to the given test
  instance, whereby closeness is evaluated using, e.g., Euclidean
  distance~\cite{aha1991instance}.
\item \textbf{Support Vector Machine} ({\svm}): it is a supervised
  learning algorithm used for both classification and regression
  tasks. Roughly speaking, the basic training principle of {\svm}s is
  finding an optimal linear hyperplane such that the expected
  classification error for (unseen) test patterns is minimized. We
  refer the reader to~\cite{cortes1995support} for further details.
\end{itemize}
The rationale of our choice is twofold. On the one hand, the selected
classifiers are ``orthogonal'', i.e., they build on different
inductive biases in the computation of their classification
hypotheses, since their classification algorithms are based on very
different approaches. On the other hand, building \measp on top of
different classifiers allows to draw conclusions about both the
robustness of our approach, and the proper design of our testing set.
Indeed, as shown in Section~\ref{sec:exp}, performance is positive
for each classification method.

As mentioned in Section~\ref{sec:multiasp}, in order to train the
classifiers, we have to select a pool of instances for training
purpose, called the training set. Concerning such selection, our aim is
twofold. On the one hand, we want to compose a training set in order to
get a robust model; while, on the other hand, we want to test the
generalization performance of \measp also on instances belonging to
benchmarks not ``covered'' by the training set.

\begin{figure*}[t]
\centering
\subfigure[Whole dataset]{\includegraphics[scale=0.36,angle=0]{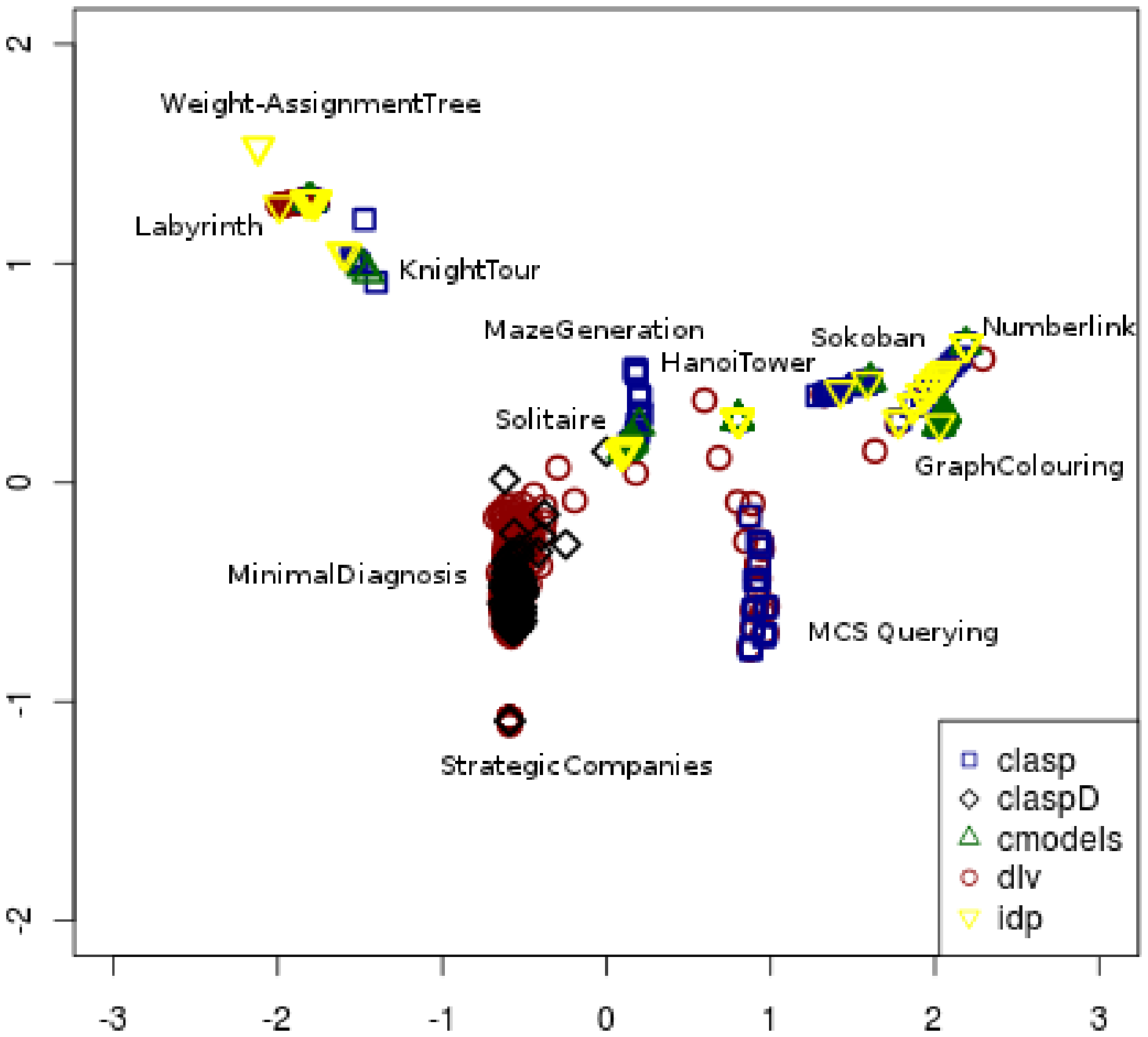}\label{fig:pca-all}}
\subfigure[\TSU]{\includegraphics[scale=0.36,angle=0]{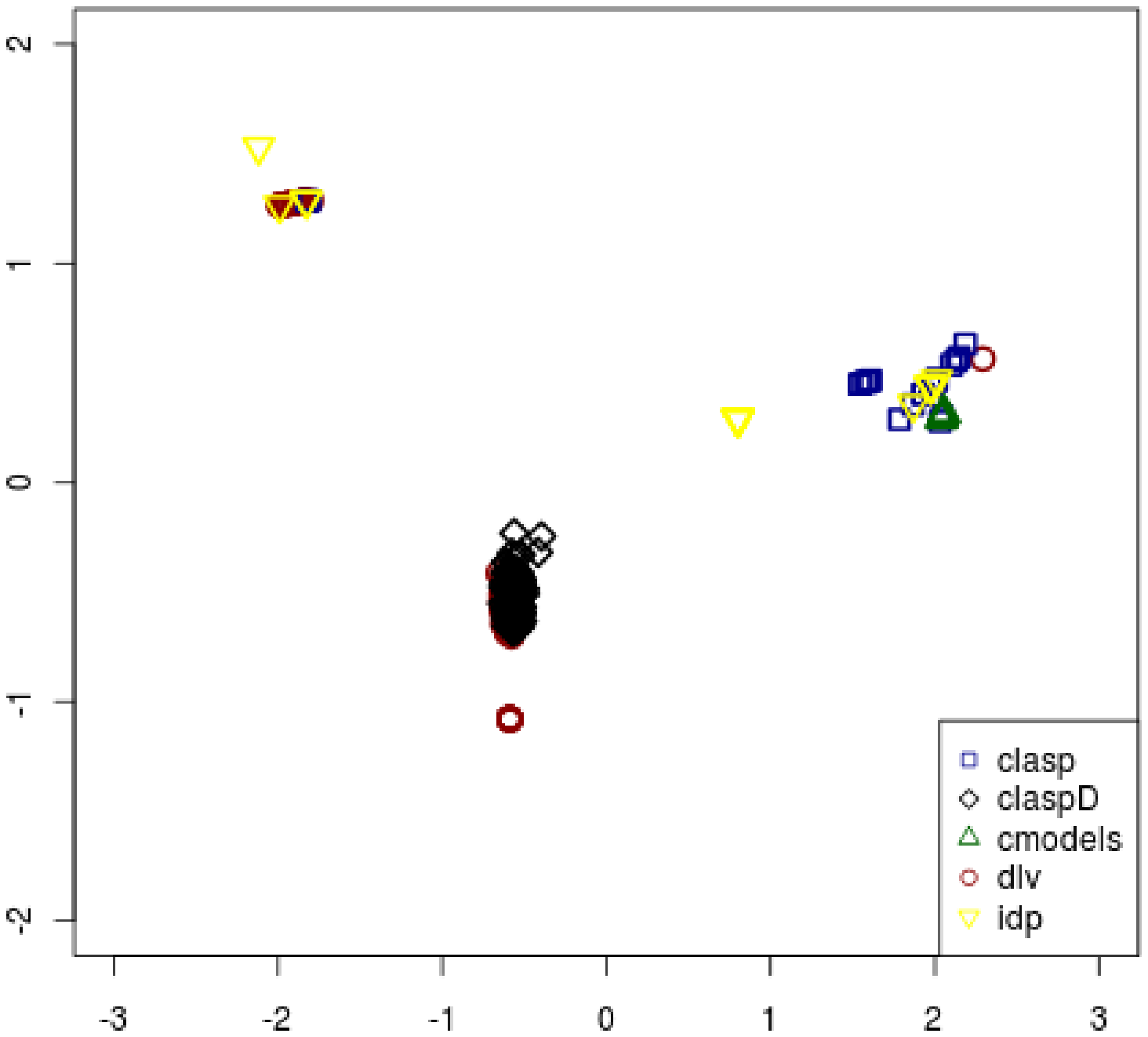}\label{fig:pca-tsu}}\\
\subfigure[\TS{S1}]{\includegraphics[scale=0.36,angle=0]{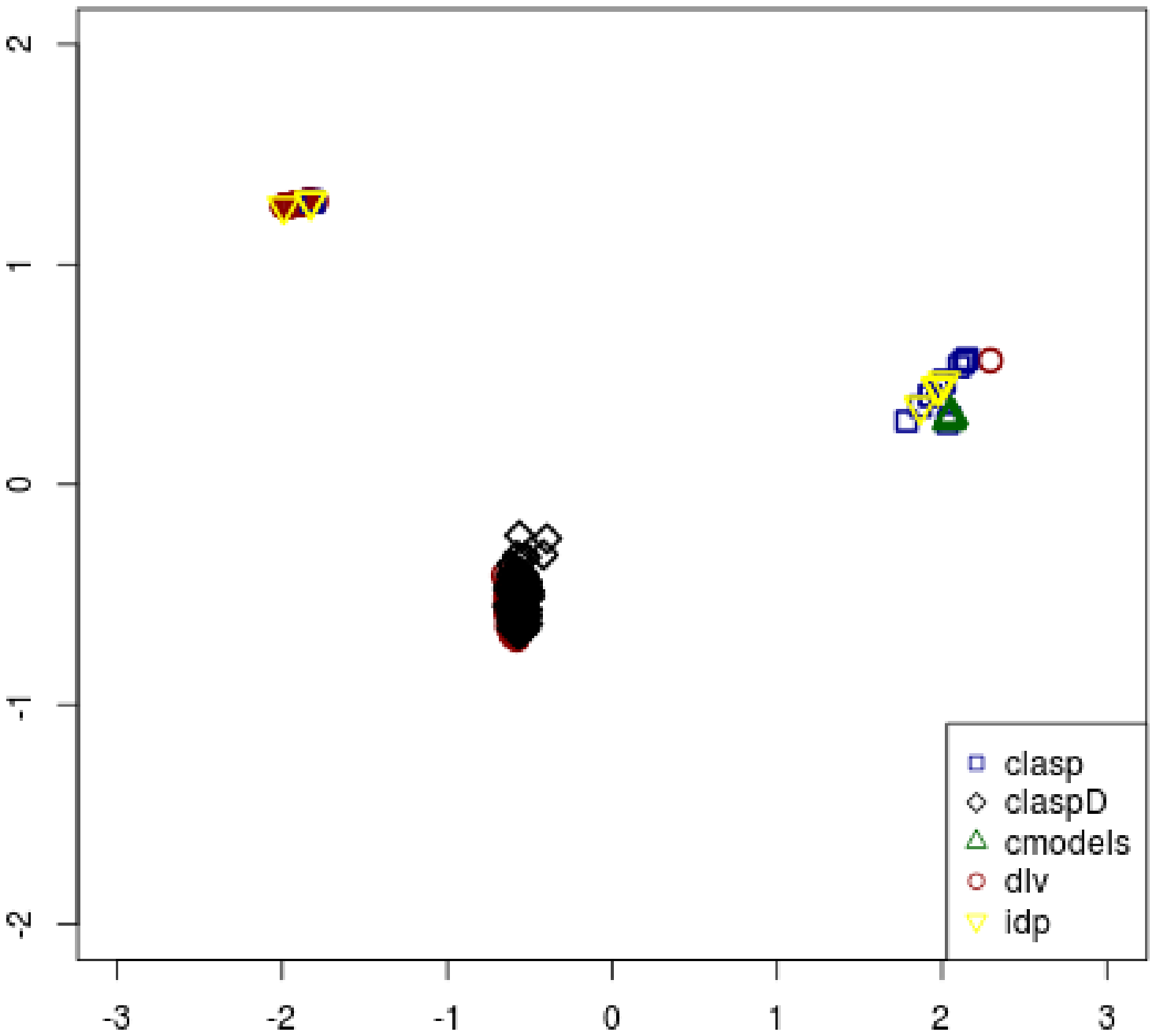}\label{fig:pca-s1}}
\subfigure[\TS {S2}]{\includegraphics[scale=0.36,angle=0]{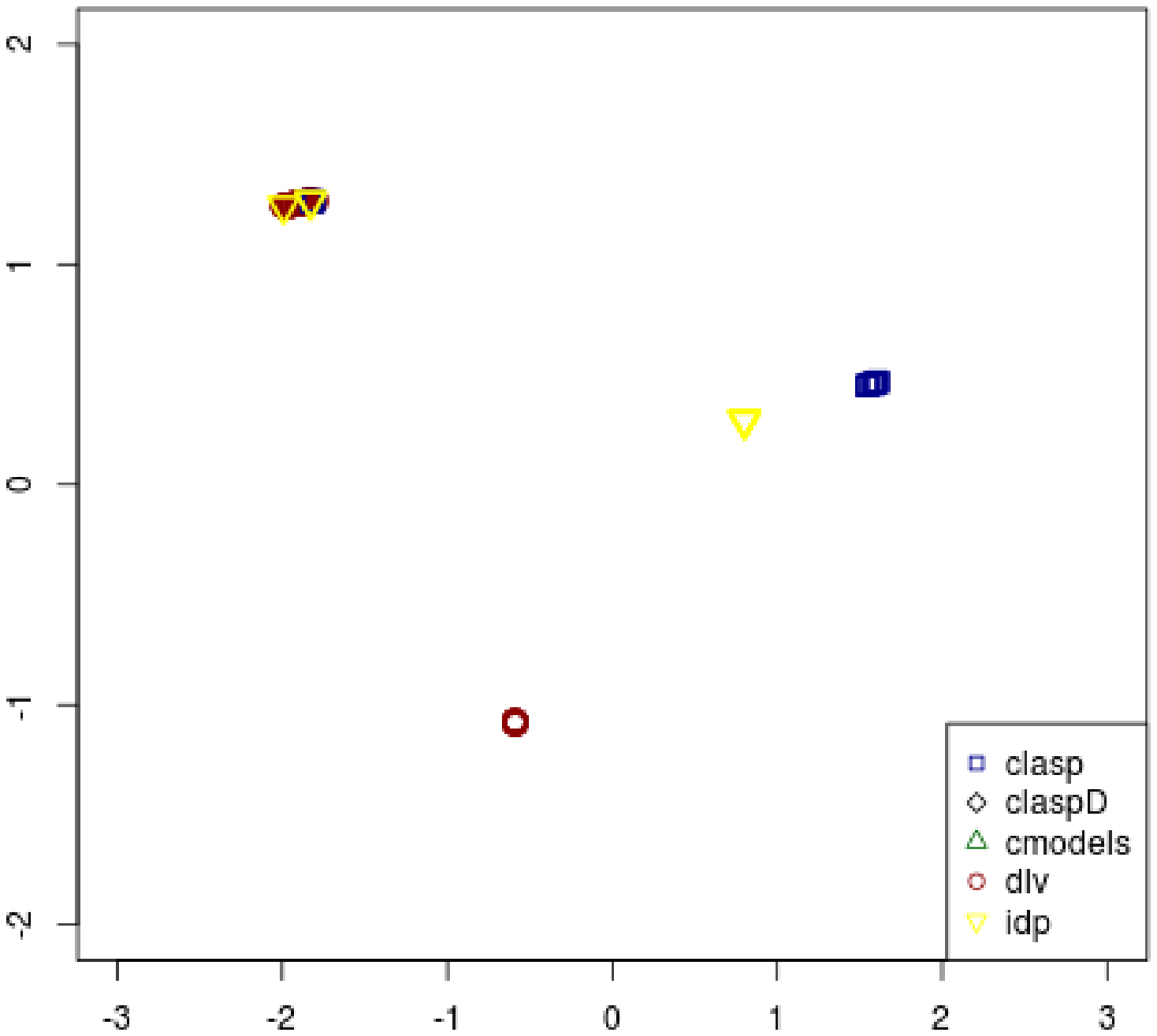}\label{fig:pca-s2}}\\
\caption{\small Training set coverage: two-dimensional space projection of (a) the whole
  dataset, (b) \TSU , (c) \TS{S1}, and (d) \TS{S2}.}\label{fig:pca}
\end{figure*}

As result of the considerations above, we designed three training sets. 
The first one --\TSU in the following-- is composed of
the 320 instances uniquely solved by the pool of engines selected
in Section~\ref{sub:solver}, i.e. such that only one engine, among the ones selected, solves each instance (without taking into account the
instances involved in the competition). The rationale of this choice is to try to
``mask'' noisy information during model training to obtain a robust model.
The remaining training sets are subsets of \TSU, and they are composed
of instances uniquely solved considering only the ones belonging 
to the problems listed in the following:
\begin{itemize}
\item[-] \TS{S1}: 297 instances uniquely solved considering: \\
  \phantom{aa}\texttt{GraphColouring}, \texttt{Numberlink}, \texttt{Labyrinth}, 
  \texttt{MinimalDiagnosis}.
\item[-] \TS{S2}: 59 instances uniquely solved considering:\\ 
  \phantom{aa}\texttt{SokobanDecision}, \texttt{HanoiTower}, \texttt{Labyrinth},
  \texttt{StrategicCompanies}.
\end{itemize}
Note that both \TS{S1} and \TS{S2} contain one distinct \BNP problem
to ensure a minimum coverage of this class of problems.
The rationale of these additional training sets is thus to test our
method on "unseen" problems, i.e. on instances coming from domains
that were not used for training: a "good" machine learning method
should generalize (to some degrees) and obtain good results also in
such setting. In this view, both training sets are composed of
instances coming from a limited number, i.e., 4 out of 14, of
problems. Moreover, \TS{S2} is also composed of a very limited number
of instances. Such setting will further challenge {\measp} to
understand what is the point in which we can have degradation in
performance: we will see that, while it is true that \TS{S2} is a
challenging situation in which performance decreases, even in this
setting {\measp} has reasonable performance and performs better than
its engines and rival systems.

In order to give an idea of the coverage of our training sets and outline differences among them,
we depict in Figure~\ref{fig:pca} the coverage of: the whole available dataset (Fig.~\ref{fig:pca-all}), 
\TSU (Fig.~\ref{fig:pca-tsu}), and its subsets \TS{S1} (Fig.~\ref{fig:pca-s1}) and \TS{S2} (Fig.~\ref{fig:pca-s2}).
In particular, the plots report a two-dimensional projection obtained by means of a
principal components analysis (PCA), and considering only the first
two principal components (PC). The $x$-axis and the $y$-axis in the
plots are the first and the second PCs, respectively. Each point in
the plots is labeled by the best solver on the related instance. In
Figure~\ref{fig:pca-all} we add a label denoting the benchmark name of the
depicted instances, in order to give a hint about the ``location'' of
each benchmark. From the picture is clear that \TS{S1} 
covers less space that \TSU, which in turn covers a 
subset of the whole set of instances. Clearly, \TS{S2},
which is the smallest set of instances, has a very limited coverage (see Fig.~\ref{fig:pca-s2}).

Considering the classification algorithms listed above,%
\footnote{For all algorithms but {\apc}, we use the 
the tool \textsc{rapidminer}~\cite{mierswa2006yale}.}
we trained the classifiers and we assessed their accuracy. 
Referring to the notation introduced in Section~\ref{sec:multiasp}, 
even assuming that a training set is sufficient to learn $f$, it is still the case that
different sets may yield a different $f$. The problem is that the
resulting trained classifier may underfit the unknown pattern --i.e.,
its prediction is wrong-- or overfit --i.e., be very accurate only
when the input pattern is in the training set.  Both underfitting and
overfitting lead to poor \emph{generalization} performance, i.e., $c$
fails to predict $f(\underline{x}^*)$ when $\underline{x}^* \neq
\underline{x}$.  However, statistical techniques can provide
reasonable estimates of the generalization error. In order to test the
generalization performance, we use a technique known as
\emph{stratified 10-times 10-fold cross validation} to estimate the
generalization in terms of {\it accuracy}, i.e., the total amount of
correct predictions with respect to the total amount of
patterns. Given a training set $(X, Y)$, we partition $X$ in subsets
$X_i$ with $i \in \{1, \ldots 10\}$ such that $X = \bigcup_{i=1}^{10}
X_i$ and $X_i \cap X_j = \emptyset$ whenever $i \neq j$; we then train
$c_{(i)}$ on the patterns $X_{(i)} = X \setminus X_i$ and
corresponding labels $Y_{(i)}$. We repeat the process 10 times, to
yield 10 different $c$ and we obtain the global accuracy estimate.

We report an accuracy greater than 92\% for each classification
algorithm trained on \TSU, while concerning the remaining
training sets, just for the sake of completeness we report an average
85\% as accuracy result.  The main reason for this result is that the
training sets different from \TSU are composed of a smaller
number of instances with respect to \TSU, thus the
classification algorithms are not able to generalize with the same
accuracy. This result is not surprising, also considering the plots in
Figure~\ref{fig:pca} and, as we will see in the experimental section,
this will influence the performance of \measp.

\section{Performance Analysis}\label{sec:exp}

In this section we present the results of the analysis we have performed. 
We consider different combinations of training and test sets, 
where the training sets are the ones introduced in
Section~\ref{sec:main}, and the test set ranges over 
the 3rd ASP Competition ground instances. 
In particular, the first (resp. second) experiment has \TSU as
training set, and the successfully grounded instances
evaluated (resp. submitted) to the 3rd ASP Competition as test set: 
the goal of this analysis is to test the {\sl efficiency} of our approach on all
the evaluated (resp. submitted) instances when the model is trained on
the whole space of the uniquely solved instances. The third experiment
considers \TS{S1} and \TS{S2} as training sets, and all the successfully
grounded instances submitted to the competition as test set: in this
case, given that the models are not trained on all the space of the
uniquely solved instances, but on a portion, and that the test set
contains ``unseen'' problems (i.e., belonging to domains that were left unknown during training), 
the goal is to test, in particular, the {\sl robustness} of our approach. 
We devoted one subsection to each of these experiments, where we 
compare {\measp} to its component engines. 
In detail, for each experiment the results are reported in a table structured as
follows: the first column reports the name of the solver and (when needed)
its inductive model in a subcolumn, where the considered inductive models are denoted by \MOD{\TSU} , \MOD{S1} and \MOD{S2},
corresponding to the test sets \TSU, \TS{S1} and \TS{S2} introduced before, respectively; the second and
third columns report the result of each solver on {\NP} and {\BNP}
classes, respectively, in terms of the number of solved instances
within the time limit and sum of their solving times (a sub-column is
devoted to each of these numbers, which are ``--'' if the related
solver was not among the selected engines).  We report the results
obtained by running {\measp} with the six classification methods
introduced in Section~\ref{subsec:classalg}, and their related
inductive models. 
In particular, \measpC{\textsc{c}}indicates {\measp} 
employing the classification method \textsc{c} $\in\{$\apc, \furia, \jqo, \mlr, \nn, \svm$\}$.
We also report the component engines employed by \measp 
on each class as explained in Section~\ref{sub:solver}, and
as reference {\sc sota}, which is the ideal multi-engine solver
(considering the engines employed).

An additional subsection summarizes results and compares
{\measp} with state-of-the-art solvers that won the 3rd ASP Competition.

We remind the reader that the compared engines were run on all the 1425 instances 
grounded in less than 600s, whereas the instances on which {\measp} was run
are limited to the ones for which we were able to compute all features (i.e., 1371 instances), 
and the timings for multi-engine systems include both the time spent 
for extracting the features from the ground instances, and the time spent by the classifier.

\begin{table}[t!]
\setlength{\tabcolsep}{1.9\tabcolsep}
\begin{center}
\begin{tabular*}{0.80\columnwidth}{lc||r|r||r|r}
\multicolumn{2}{c||}{\bf Solver} & \multicolumn{2}{|c||}{\bf\NP} & \multicolumn{2}{|c}{\bf\BNP}\\
\cline{2-6}
&\multicolumn{1}{|c||}{Ind. Model}& \multicolumn{1}{c|}{\#Solved} & \multicolumn{1}{c||}{Time} & \multicolumn{1}{c|}{\#Solved} & \multicolumn{1}{c}{Time} \\ 
\cline{1-6}
\cline{1-6}
\multicolumn{2}{l||}{ \clasp} & 60 & 5132.45 & -- & -- \\ 
\multicolumn{2}{l||}{\claspD} & -- & --  & 13 & 2344.00 \\ 
\multicolumn{2}{l||}{\cmodels} & 56 & 5092.43 & 9 & 2079.79 \\ 
\multicolumn{2}{l||}{\dlv} & 37 & 1682.76 & 15 & 1359.71 \\ 
\multicolumn{2}{l||}{\idp} & 61 & 5010.79 & -- & -- \\ \cline{1-6} 
\multicolumn{1}{l|}{\measpC{\apc}} & \MOD{\TSU}  & 63 & 5531.68 & 15 & 3286.28 \\ 
\multicolumn{1}{l|}{\measpC{\furia}} & \MOD{\TSU}  & 63 & 5244.73 & 15 & 3187.73 \\ 
\multicolumn{1}{l|}{\measpC{\jqo}} & \MOD{\TSU}  & 68 & 5873.25 & 15 & 3187.73 \\ 
\multicolumn{1}{l|}{\measpC{\mlr}} & \MOD{\TSU}  & 65 & 5738.79 & 15 & 3187.57 \\ 
\multicolumn{1}{l|}{\measpC{\nn}} & \MOD{\TSU}  & 66 & 4854.78 & 15 & 3187.31 \\ 
\multicolumn{1}{l|}{\measpC{\svm}} & \MOD{\TSU}  & 60 & 4830.70 & 15 & 2308.60 \\ \cline{1-6}
\multicolumn{2}{l||}{\textsc{sota}} & 71 & 5403.54 & 15 & 1221.01 \\ 
\cline{1-6}
\end{tabular*}
\caption{\small Results of the various solvers on the grounded instances
  evaluated at the 3rd ASP Competition. {\measp} has been trained on
the \TSU training set. 
}\label{tab:exp1}
\end{center}
\end{table}

\subsection{Efficiency on Instances Evaluated at the Competition}
\label{subsec:exp1}

In the first experiment we consider \TSU introduced in Section~\ref{sec:main} as training set, 
and as
test set all the instances evaluated at the 3rd ASP Competition 
(a total of 88 instances).  
Results are shown in Table~\ref{tab:exp1}. We can see
that, on problems of the {\NP} class, \measpC{\jqo} solves 
the highest number of instances, 7 more than \idp and 8 more than {\clasp}.
Note also that \measpC{\svm} (our worst performing version) 
is basically on par with \clasp (with 60 solved instances) 
and is very close to \idp (with 61 solved instances).
Nonetheless, 5 out of 6 classification methods lead {\measp} to have
better performance than each of its engines. On the
{\BNP} problems, instead, all versions of {\measp} and {\dlv} solve 15
instances ({\dlv} having best mean CPU time), followed by {\claspD}
and {\cmodels}, which solve 13 and 9 instances, respectively. Among the {\measp}
versions, \measpC{\jqo} is, in sum, the solver that solves
the highest number of instances: here it is very interesting to note
that its performance is very close to the {\sc sota} solver (solving
only 3 instances less) which, we
remind, has the ideal performance that we could expect in these
instances with these engines. 

\subsection{Efficiency on Instances Submitted to the Competition}

In the second experiment we consider the \TSU training set 
(as for the previous experiment), and the test set is composed of all
successfully grounded instances submitted to the 3rd ASP Competition.
The results are now shown in Table~\ref{tab:exp2}. Note here that in both {\NP} and {\BNP} classes, all {\measp}
versions solve more instances (or in shorter time in one case) than the
component engines: in particular, in the {\NP} class,
\measpC{\apc} solves the highest number of instances, 52 more than
{\clasp}, which is the best engine in this class, %
while in the {\BNP} class \measpC{\mlr} solves 519 instances and 
three {\measp} versions solve
518 instances, i.e., 86 and 85 more instances than {\claspD}, respectively, which is the
engine that solves more instances in the {\BNP} class. 
Also in this case \measpC{\svm} solves less instances than other \measp versions;
nonetheless, \measpC{\svm} can solve as much \NP instances as \clasp, 
and is effective on \BNP, where it is one of the versions 
that can solve 518 instances.

As far as the comparison with the {\sc sota} solver is concerned, 
the best {\measp} version, i.e., \measpC{\apc}  solves, in sum, 
only 23 out of 1036 instances less than the {\sc sota} solver, 
mostly from the {\NP} class.

\begin{table}[t!]
\setlength{\tabcolsep}{1.9\tabcolsep}
\begin{center}
\begin{tabular*}{0.80\columnwidth}{lc||r|r||r|r}
\multicolumn{2}{c||}{\bf Solver} & \multicolumn{2}{|c||}{\bf\NP} & \multicolumn{2}{|c}{\bf\BNP}\\
\cline{2-6}
&\multicolumn{1}{|c||}{Ind. Model}& \multicolumn{1}{c|}{\#Solved} & \multicolumn{1}{c||}{Time} & \multicolumn{1}{c|}{\#Solved} & \multicolumn{1}{c}{Time} \\ 
\cline{1-6}
\cline{1-6}
\multicolumn{2}{l||}{\clasp} & 445 & 47096.14 & -- & -- \\ 
\multicolumn{2}{l||}{\claspD} & -- & --  & 433 & 52029.74 \\ 
\multicolumn{2}{l||}{\cmodels} & 333 & 40357.30 & 270 & 38654.29 \\ 
\multicolumn{2}{l||}{\dlv} & 241 & 21678.46 & 364 & 9150.47 \\ 
\multicolumn{2}{l||}{\idp} & 419 & 37582.47 & -- & -- \\ \cline{1-6} 
\multicolumn{1}{l|}{\measpC{\apc}} &\MOD{\TSU}  & 497 & 55334.15 & 516 & 60537.67 \\ 
\multicolumn{1}{l|}{\measpC{\furia}} &\MOD{\TSU}  & 480 & 48563.26 & 518 & 60009.23 \\ 
\multicolumn{1}{l|}{\measpC{\jqo}} &\MOD{\TSU}  & 490 & 49564.19 & 510 & 59922.86 \\ 
\multicolumn{1}{l|}{\measpC{\mlr}} &\MOD{\TSU}  & 489 & 49569.77 & 519 & 58287.31\\ 
\multicolumn{1}{l|}{\measpC{\nn}} &\MOD{\TSU}  & 490 & 46780.31 & 518 & 55043.39 \\ 
\multicolumn{1}{l|}{\measpC{\svm}} &\MOD{\TSU}  & 445 & 40917.70 & 518 & 52553.84 \\ \cline{1-6} 
\multicolumn{2}{l||}{\textsc{sota}} & 516 & 39857.76  & 520 & 24300.82 \\
\cline{1-6}
\end{tabular*}
\caption{\small Results of the various solvers on the grounded instances
submitted to the 3rd ASP Competition. {\measp} has been trained on
the \TSU training set.}\label{tab:exp2}
\end{center}
\end{table}

\begin{figure*}[t]
\centering
\includegraphics[scale=0.55,angle=0]{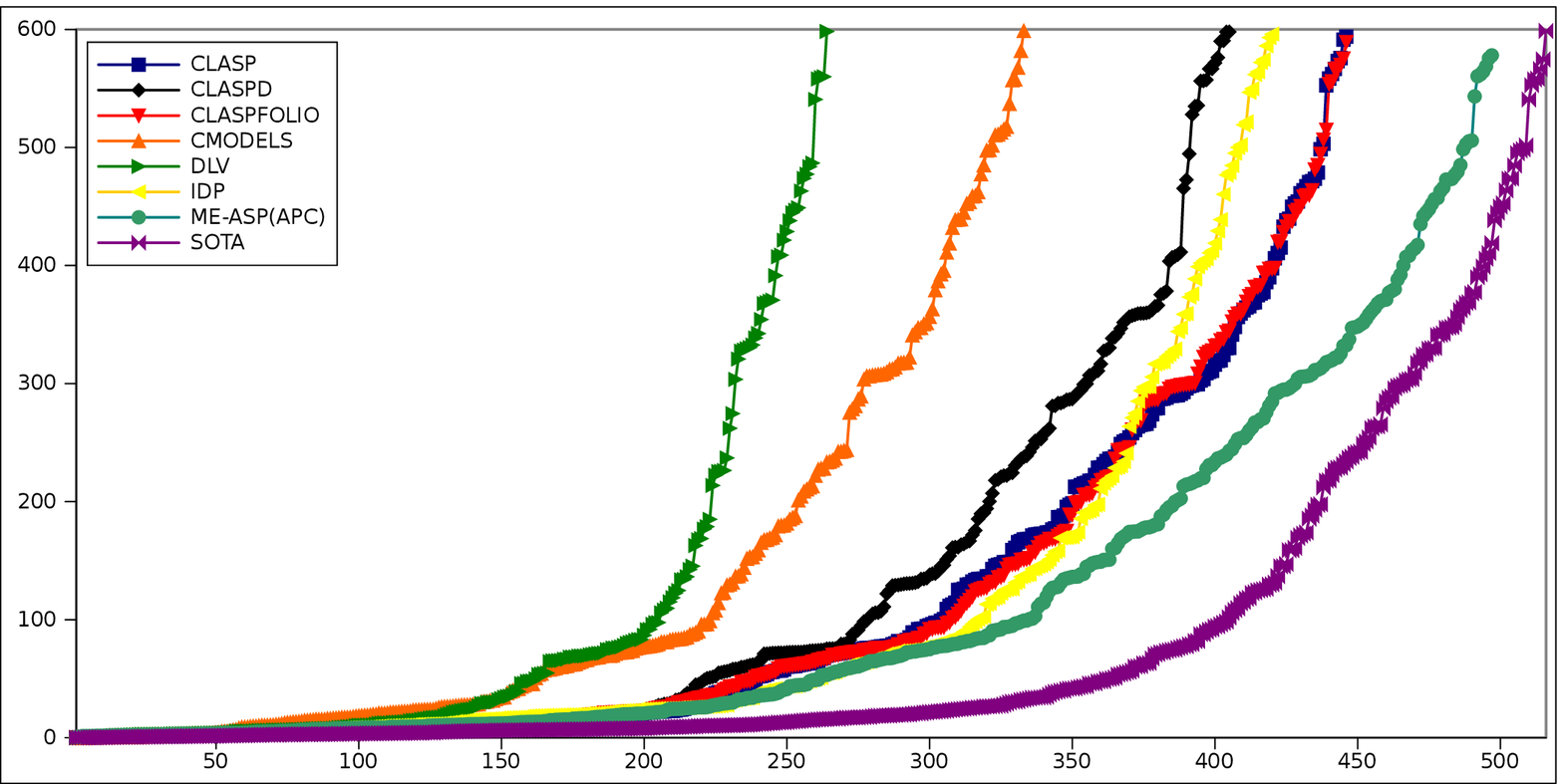}
\caption{Results of {\claspf}, {\measp} engines, {\measpC{\apc}} (trained on \TS) and {\sota} on the {\NP} instances submitted to the competition. }\label{fig:cactus}
\end{figure*}

In order to give a different look at the magnitude of improvements of
our approach in this experiment, whose test set we remind is a
super-set of the one in Section~\ref{subsec:exp1}, in
Fig~\ref{fig:cactus} we present the results of {\measpC{\apc}}, its
engines, {\claspf} and {\sc sota} on {\NP} instances in a
cumulative way as customary in, e.g., Max-SAT and ASP
Competitions. The $x$-axis reports a CPU time, while the $y$-axis
indicates the number of instances solved within a certain CPU time.

Results clearly show that {\measpC{\apc}} performs better, in terms of
total number of instances solved, than its engines {\clasp}, {\claspD} and
{\claspf}; also, {\measpC{\apc}} it is very close to the {\sota}. Looking more in
details at the figure, we can note that, along the $x$-axis the
distance of {\measpC{\apc}} w.r.t. the {\sota} decreases: this is due,
for a small portion of instances (given that we have seen that these two steps are efficient), 
to the time spent to compute features and on classification, and to
the fact that we may not always predict the best engine to run. 
The convergence of {\measpC{\apc}} toward {\sota} confirms that, even if we 
may sometimes miss to predict the best engine, most of the time we predict an
engine that allows to solve the instance within the time limit.

\subsection{Robustness on Instances Submitted to the Competition}

In this experiment, we use the two smaller training sets \TS{S1} and \TS{S2} 
introduced in Section~\ref{sec:main},
while the same test set as that of previous experiment. The
rationale of this last experiment is to test the robustness of our approach on ``unseen''
problems, i.e., in a situation where the test set does not contain any instance
from some problems. 
Note that \TS{S1} contains 297 uniquely solved instances, covering 4 domains out of 14;
and \TS{S2} is very small, since it contains only 59 instances belonging to 4 domains.
We can thus expect this experiment to be particularly
challenging for our multi-engine approach.  
Results are presented in Table~\ref{tab:exp3}, from which 
it is clear that {\measpC{\apc}}  trained on \TS{S1} performs 
better that the other alternatives and solves 46 instances 
more than {\clasp} in the {\NP} class, 
and 11 instances more than {\claspD} in the {\BNP} class ({\clasp} and {\claspD} 
being the best engines in \NP and \BNP classes, respectively). 
As expected, if we compare the results with the ones obtained with the larger training set \TSU, 
we note a general performance degradation. In particular, 
the performance now is less close to the {\sc sota} solver, which
solves in total 40 more instances than the best \measp version trained on \TS{S1}, with additional unsolved instances coming mainly from the {\BNP} class in this case.
This can be explained considering that \TS{S1} does not contain instances from
the Strategic Companies problem, and, thus, it is not always able to select \dlv on these instances
where \dlv is often a better choice than \claspD.
However, \measp can solve also in this case far more instances than all the engines, 
demonstrating a robust performance.

These findings are confirmed when the very small test set \TS{S2} is considered.
In this very challenging setting there are still \measp versions 
that can solve more instances than the component engines.

\begin{table}[t!]
\setlength{\tabcolsep}{1.9\tabcolsep}
\begin{center}
\begin{tabular*}{0.80\columnwidth}{lc||r|r||r|r}
\multicolumn{2}{c||}{\bf Solver} & \multicolumn{2}{|c||}{\bf\NP} & \multicolumn{2}{|c}{\bf\BNP}\\
\cline{2-6}
&\multicolumn{1}{|c||}{Ind. Model}& \multicolumn{1}{c|}{\#Solved} & \multicolumn{1}{c||}{Time} & \multicolumn{1}{c|}{\#Solved} & \multicolumn{1}{c}{Time} \\ 
\cline{1-6}
\cline{1-6}
\multicolumn{2}{l||}{\clasp} & 445 & 47096.14 & -- & -- \\
\multicolumn{2}{l||}{\claspD} & -- & --  & 433 & 52029.74 \\
\multicolumn{2}{l||}{\cmodels} & 333 & 40357.30 & 270 & 38654.29 \\ 
\multicolumn{2}{l||}{\dlv} & 241 & 21678.46 & 364 & 9150.47 \\ 
\multicolumn{2}{l||}{\idp} & 419 & 37582.47 & -- & -- \\ \cline{1-6} 
\multicolumn{1}{l|}{\measpC{\apc}} & \MOD{S1} & 491 & 54126.87 & 505 & 56250.96 \\
\multicolumn{1}{l|}{\measpC{\furia}} & \MOD{S1} & 479 & 49226.42 & 507 & 55777.67 \\
\multicolumn{1}{l|}{\measpC{\jqo}} & \MOD{S1} & 477 & 46746.65 & 507 & 55777.67 \\
\multicolumn{1}{l|}{\measpC{\mlr}} & \MOD{S1} & 471 & 48404.11 & 507 & 52499.83 \\ 
\multicolumn{1}{l|}{\measpC{\nn}} & \MOD{S1} & 476 & 47627.06 & 507 & 49418.67 \\ 
\multicolumn{1}{l|}{\measpC{\svm}} & \MOD{S1} & 459 & 38686.16 & 507 & 51462.13 \\ \cline{1-6}
\multicolumn{1}{l|}{\measpC{\apc}} & \MOD{S2} & 445 & 48290.97 & 433 & 53268.62 \\
\multicolumn{1}{l|}{\measpC{\furia}} & \MOD{S2} & 414 & 37902.37 & 363 & 10542.85 \\
\multicolumn{1}{l|}{\measpC{\jqo}} & \MOD {S2} & 487 & 51187.66 & 431 & 57393.61 \\
\multicolumn{1}{l|}{\measpC{\mlr}} & \MOD{S2} & 460 & 42385.66 & 363 & 10542.01 \\ 
\multicolumn{1}{l|}{\measpC{\nn}} & \MOD{S2} & 487 & 48889.21 & 363 & 10547.81 \\
\multicolumn{1}{l|}{\measpC{\svm}} & \MOD{S2} & 319 & 32162.37 & 364 & 10543.00 \\ \cline{1-6} 
\multicolumn{2}{l||}{\textsc{sota}} & 516 & 39857.76  & 520 & 24300.82 \\ 
\cline{1-6}
\end{tabular*}
\caption{\small Results of the various solvers on the grounded instances
  submitted to the 3rd ASP Competition. {\measp} has been trained on training sets 
\TS{S1} and \TS{S2}. }\label{tab:exp3}
\end{center}
\end{table}

\begin{figure*}[t]
\centering
\subfigure[Inductive model \MOD{\TSU} ]{\includegraphics[scale=0.45,angle=0]{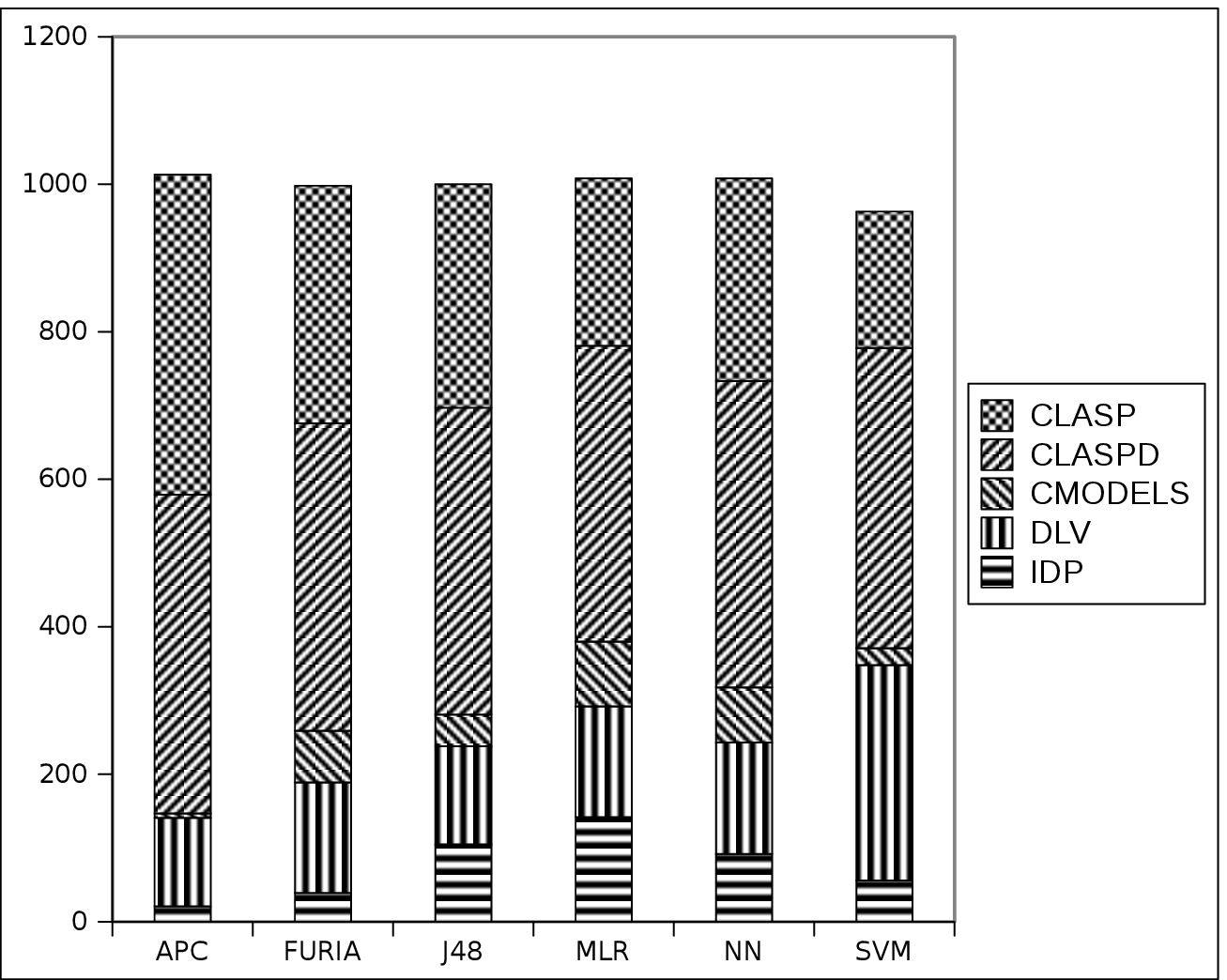}\label{fig:callunique}}\\
\subfigure[Inductive model \MOD{S1}]{\includegraphics[scale=0.46,angle=0]{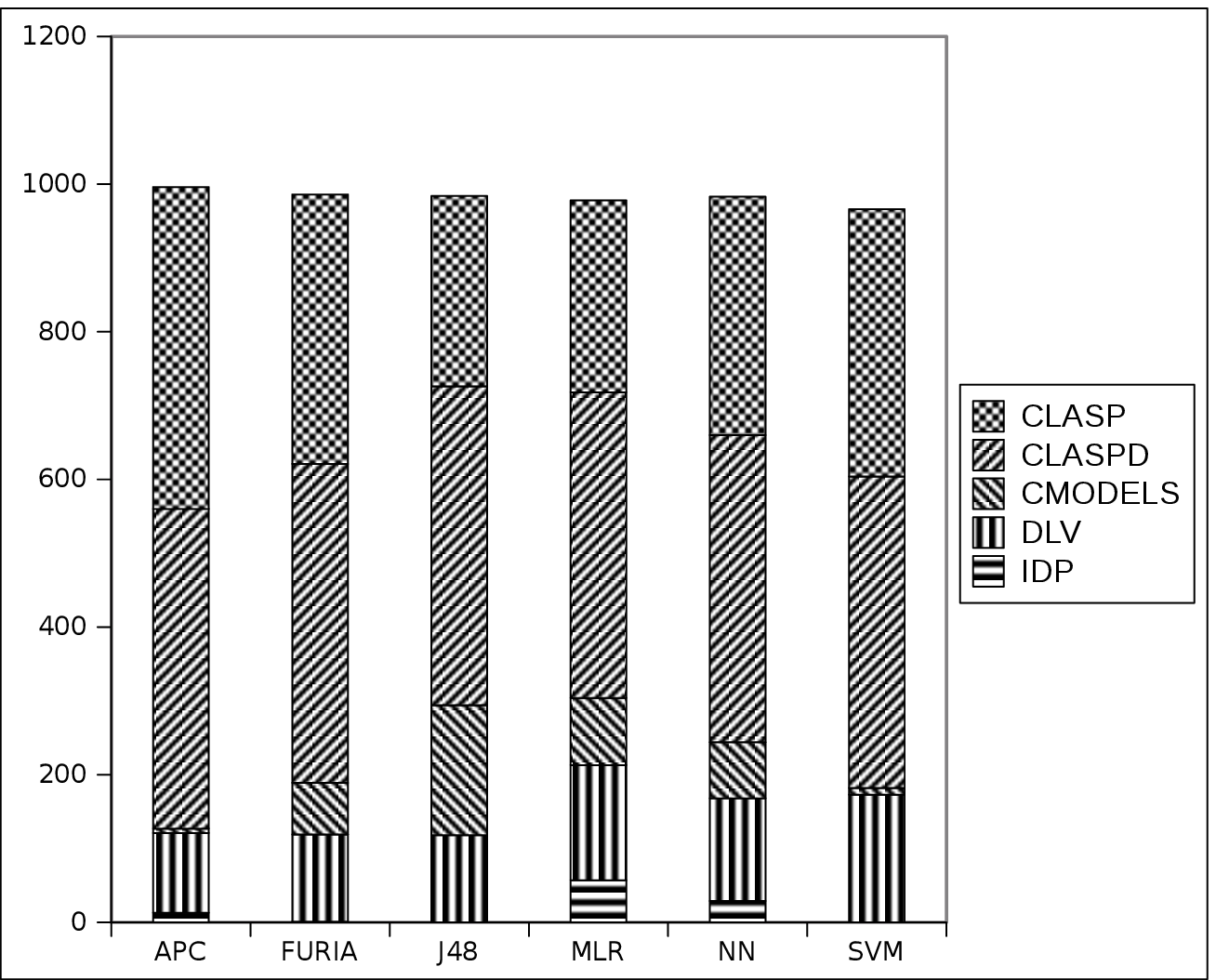}\label{fig:callS1}}\\
\subfigure[Inductive model \MOD{S2}]{\includegraphics[scale=0.46,angle=0]{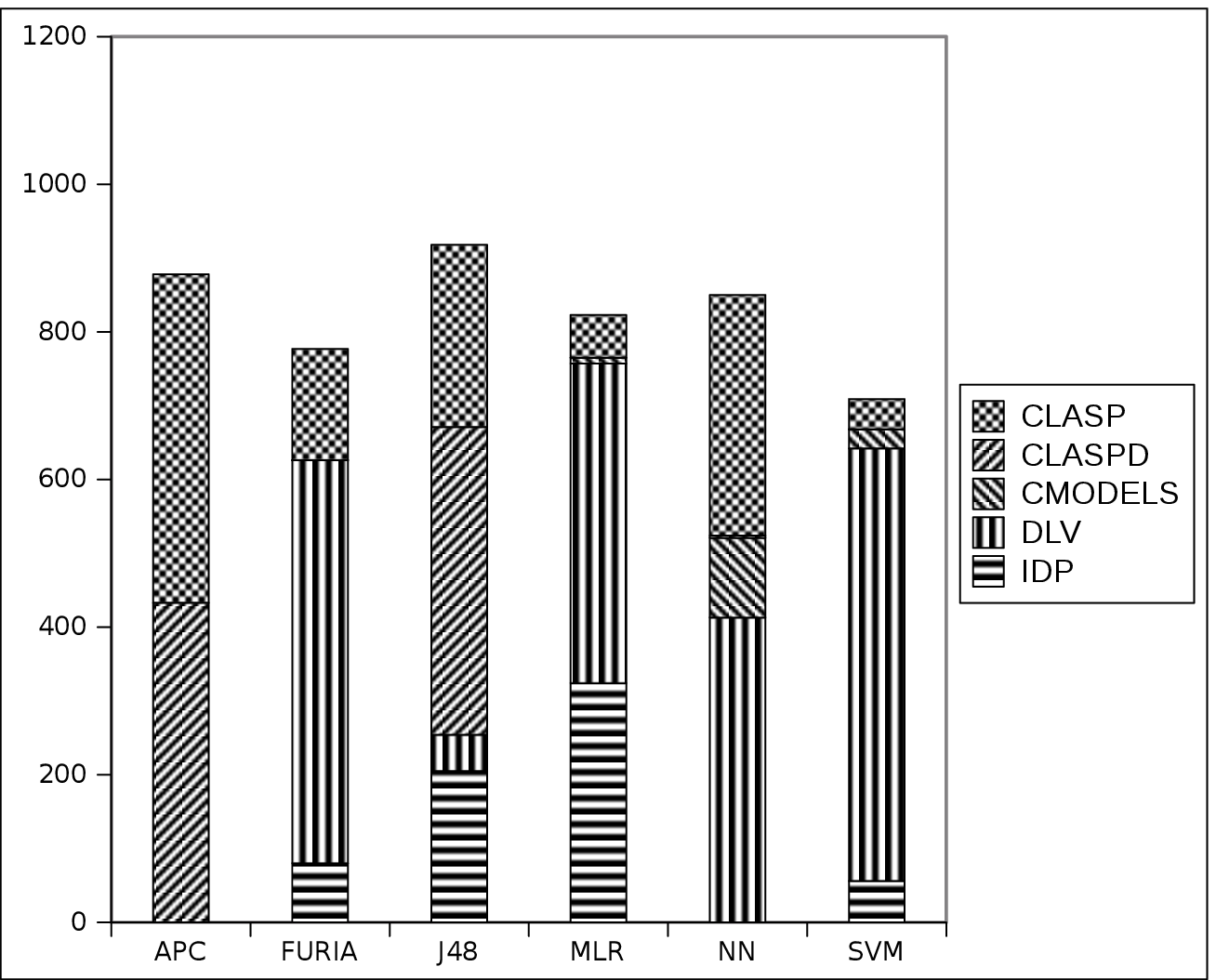}\label{fig:callS2}}\\
\caption{Number of calls to the component engines of the various versions of \measp on the instances {\em submitted} to the 3rd ASP Competition.}\label{fig:calls}
\end{figure*}

\subsection{Discussion and Comparison to the State of the Art}\label{sub:discussion}

Summing up the three experiments, %
it is clear that {\measp} has a very robust and efficient performance: it
often can solve (many) more instances than its engines, %
even considering the single {\NP} and {\BNP} classes. 

We also report that all versions of
{\measp} have reasonable performance, so --from a machine learning
point of view-- we can conclude that, on the one hand, 
the set of cheap-to-compute features that we selected 
is representative (i.e., they allow to both analyze a significant number of instances and
drive the selection of an appropriate engine)
independently from the classification method employed. 
On the other hand, the robustness of our inductive models 
let us conclude that we made an appropriate design of our training set \TSU.

Additional observations can be drawn by looking at Figure~\ref{fig:calls}, 
where three plots are depicted, one for each inductive model, 
showing the number of calls to the internal engines for each variant of \measp. 
In particular, by looking at Figure~\ref{fig:callunique}, we can conclude that also the selection of the
engines was fair. Indeed, all of them were employed in a significant number of cases and,
as one would expect, the engines that solved a larger number of instances in the 3rd ASP Competition 
(i.e., \clasp and \claspD) are called more often. Nonetheless, 
the ability of exploiting all solvers from the pool made a difference in performance, 
e.g., looking at Figure~\ref{fig:callunique} one can note 
that our best version \measpC{\apc} exploits all engines, and 
it is very close to the ideal performance of \sota. 
It is worth noting that the \measp versions that select \dlv 
more often (note that \dlv solves uniquely a high number of StrategicCompanies instances) performed better on \BNP.
Note also that Figure~\ref{fig:calls} allows to explain
the performance of \measpC{\svm}, which often differs from the other methods; 
indeed, this version often prefers \dlv  over the other engines also on \NP instances.  
Despite choosing \dlv is often decisive on \BNP, it is not always a good choice on \NP\ as well. 
As a consequence \measpC{\svm} is always very fast on \BNP 
but does not show overall the same performance
of \measp equipped with other methods.

Figure~\ref{fig:callunique} also gives some additional insight concerning
the differences among our inductive models. In particular, the \measp versions trained with \TS{S2} 
(containing only StrategicCompanies in \BNP) prefer more often \dlv (see Fig.~\ref{fig:callS2}),
thus the performance is good on this class but deteriorates a bit on \NP.
Concerning \TS{S1} (see Fig.~\ref{fig:callS1}), we note that \idp is less exploited than in the other cases, even by {\measpC{\mlr}} which is the alternative
that chooses \idp more often: this is probably due to the minor coverage of this training set on \NP. On the overall, as we would expect, the number of calls for \measp trained 
with \TSU is more balanced among the various engines, 
than for \measp trained with the smaller training sets.

\medskip

We have seen that  \measp almost always can solve more instances than its component engines. 
One might wonder how it compares with the state-of-the-art ASP implementations.
Table~\ref{tab:sota} summarizes the performance of \claspD and \claspf
(the overall winner, and the fastest solver in the \NP class
that entered the System Track of the competition, respectively), 
in terms of number of solved instances on both instance sets, 
i.e., evaluated and submitted,
and of the various versions of \measp exploiting our inductive model of choice,
obtained from the test set \TSU.

We observe that all \measp versions outperform yardstick state-of-the-art solvers considering all submitted instances.\footnote{Recall that \claspf can deal with \NP instances only.}

Concerning the comparison on the instances {\em evaluated} at the 3rd ASP Competition,
we note that all \measp versions outperform the winner of the System Track of the competition \claspD 
that could solve 65 instances, whereas \measpC{\jqo} (i.e., the best solver in this class) solves 83 instances
(and is very close to the ideal \sota solver). Even
\measpC{\svm} (i.e., the worst performing version of our system) could solve 10 instances more than \claspD; moreover, also  \measpC{\apc} is very effective here, solving 78 instances.

Concerning the comparison on the larger set of instances {\em submitted} to the 3rd ASP Competition,
the picture is similar. All \measp versions outperform \claspD, which solves 835 instances
where  the worst performing version of our system, \measpC{\svm}, solves 963 instances, and 
the best version overall \measpC{\apc} solvers 1013 instances, i.e., 178 instances
more that the winner of the 3rd ASP System competition.
We remind that this holds even considering the most challenging settings when \measp is trained 
with \TS{S1} and \TS{S2} (see Tab.~\ref{tab:exp3}).

If we limit our attention to the instances belonging to the \NP class, 
the yardstick for comparing \measp with the state of the art is clearly \claspf.
Indeed, \claspf was the solver that  could solve more \NP instances
at the 3rd ASP Competition, and also \claspf is the state of the art portfolio system for ASP,
selecting from a pool of different \clasp configurations.

The picture that comes out from Table~\ref{tab:sota} shows that all versions of
\measp could solve more instances than \claspf, especially considering the instances {\em submitted} to the competition.
In particular, \measpC{\apc} solves 497 \NP instances, while \claspf solves 431.
Concerning the comparison on the instances {\em evaluated} at the 3rd ASP Competition,
we note that \claspf could solve 62 instances and performs similarly to, e.g.,
\measpC{\svm} (with  60 instances), and \measpC{\furia} (with  63 instances);
our best performing version (i.e., \measpC{\jqo}) could solve 68 instances, i.e., 6 instances more that \claspf (i.e., about 10\% more).

\medskip

Up to now we have compared the raw performance of \measp with out-of-the-box alternatives.
A more precise picture of the comparison between the two machine learning based approaches 
(\measp and \claspf) can be obtained by performing some additional analysis.

First of all note that the above comparison was made considering as reference
the \claspf version (trained by the Potassco team) that entered the 3rd ASP Competition. 
One might wonder what is the performance of \claspf when trained on our training set \TSU.
As will be discussed in detail in Section~\ref{sec:related}, \claspfolio exploits a different 
method for algorithm selection, thus this datum is reported here only for the sake of completeness.
We have trained \claspf on \TSU with the help of the Potassco team.%
\footnote{Following the suggestion of the Potassco team we have run 
\claspfolio (ver. 1.0.1 -- Aug, 19th 2011), since the feature extraction tool {\sc claspre} 
has been recently updated and integrated in \claspfolio.}
As a result, the performance of \claspf trained on \TSU is analogous to the one obtained by 
the \claspf  trained for the competition (i.e., it solves 59 instances from 
the {\em evaluated} set, and 433 of the {\em submitted} set).

On the other hand, one might want to analyze what would be the result of applying 
the approach to algorithm selection implemented in \measp to the setting of \claspf.
As pointed out in Section~\ref{sec:related} the multi-engine approach that 
we have followed in \measp is very flexible, and we could easily develop an ad-hoc version
of our system, that we called \meclasp, that is based on the same ``algorithms'' portfolio of \claspf.
In practice, we considered as a separate engine each of the 25 \clasp versions employed 
in \claspfolio, and we applied the same steps described in Section~\ref{sec:main} to build \meclasp.
Concerning the selection of the engines, as one might expect, many engines are overlapping
and the number of uniquely solved instances considering all the available engines was very low (we get only ten uniquely solved instances).
Thus, we applied the extended engine selection policy and we selected 5 engines, 
we trained \meclasp on \TSU, and selected a classification algorithm, in this case \nn.
(We also tried other settings with different combinations, both more and less engines,
still obtaining similar overall results).

The goal of this final experiment is to confirm the prediction power of our approach. 
The resulting picture is that \meclasp(\nn) solves 458 \NP instances, where the ideal limit that one can reach 
considering all the 25 heuristics in the portfolio is 484. This is substantially more that \claspf, solving 431 instances.
Nonetheless, \measp(\nn) (that solves 490) outperforms \meclasp(\nn). 

All in all, one can conclude that the approach introduced in this paper, combining cheap-to-compute features, 
and multinomial classification works well also when applied to a portfolio of heuristics.
On the other hand, as one might expect, the possibility to select among several different engines featuring 
(often radically different) evaluation strategies with non overlapping performance, 
gives additional advantages w.r.t. a single-engine portfolio. 
Indeed, even in presence of an ideal prediction strategy, a portfolio approach
based on variants of the same algorithm cannot achieve the same performance 
of an ideal multi-engine approach. 
This is clear observing that the \sota solver  on \NP can solve 516 instances, 
whereas the ideal performance for both \meclasp and \claspf tops at 484 instances.
The comparison of \meclasp and \measp seem to confirm that 
\measp can exploit this ideal advantage also in practice.

\begin{table}[t!] \hspace*{-2cm}
\setlength{\tabcolsep}{1.9\tabcolsep}
\begin{tabular*}{0.80\columnwidth}{lc||r|c|r||r|c|r}
\multicolumn{2}{c||}{\bf Solver} & \multicolumn{3}{|c||}{\bf Evaluated } & \multicolumn{3}{|c}{\bf Submitted }\\
\cline{2-8}
&\multicolumn{1}{|c||}{Ind. Model}& \multicolumn{1}{c|}{\NP } & \multicolumn{1}{c|}{\BNP } & \multicolumn{1}{c||}{\bf Tot.} & \multicolumn{1}{|c|}{\NP} & \multicolumn{1}{c|}{\BNP } & \multicolumn{1}{c}{\bf Tot.}\\ 
\cline{1-8}
\cline{1-8}
\multicolumn{1}{l|}{{\claspD}} & --  & 52 & 13 & 65 & 402 & 433 & 835 \\ 
\multicolumn{1}{l|}{\claspfolio}& Competition & 62 & -- & -- & 431 & -- & -- \\ 
\multicolumn{1}{l|}{\measpC{\apc}} &\MOD{\TSU}   & 63 & 15 & 78 & 497 & 516 & {\bf 1013} \\ 
\multicolumn{1}{l|}{\measpC{\furia}} &\MOD{\TSU}   & 63 & 15 & 78 & 480 & 518 & 998\\ 
\multicolumn{1}{l|}{\measpC{\jqo}} &\MOD{\TSU}   & 68 & 15 & {\bf 83} & 490 & 510 & 1000\\ 
\multicolumn{1}{l|}{\measpC{\mlr}} &\MOD{\TSU}   & 65 & 15 & 80 & 489 & 519 & 1008\\ 
\multicolumn{1}{l|}{\measpC{\nn}} &\MOD{\TSU}   & 66 & 15 & 81 & 490 & 518 & 1008\\ 
\multicolumn{1}{l|}{\measpC{\svm}} &\MOD{\TSU}  & 60 & 15 & 75 & 445 &  518 & 963\\ 
\cline{1-8}
\end{tabular*}
\caption{Comparison to the state of the art. \measp trained on training set \TSU.}
\label{tab:sota}
\end{table}

\section{Related Work}\label{sec:related}

Starting from the consideration that, on empirically hard problems,
there is rarely a ``global'' best algorithm, while it is often the
case that different algorithms perform well on different problem
instances, Rice~\citeyear{rice-76} defined the algorithm selection problem
as the problem of finding an effective algorithm
based on an abstract model of the problem at hand. Along this line,
several works have been done to tackle combinatorial problems
efficiently. In \cite{gome-selm-2001,leyton2003portfolio} it is described the
concept of ``algorithm portfolio'' as a general method for combining
existing algorithms into new ones that are unequivocally preferable to
any of the component algorithms. Most related papers to our work
are~\cite{xu-etal-2008,puli-tacc-07} for solving SAT and QSAT
problems. Both~\cite{xu-etal-2008} and~\cite{puli-tacc-07} rely on a
per-instance analysis, like the one we have performed in this paper:
in~\cite{puli-tacc-07}, which is the work closest to our, the goal is
to design a multi-engine solver, i.e. a tool that can choose among its
engines the one which is more likely to yield optimal results. 
\cite{puli-tacc-09} extends~\cite{puli-tacc-07} by introducing a self-adaptation of the
learned selection policies when the approach fails to give a good prediction.
The approach by Xu et al.~\citeyear{xu-etal-2008} has also the ability to compute
features on-line, e.g., by running a solver for an allotted amount of
time and looking ``internally'' to solver statistics, with the option
of changing the solver on-line: this is a per-instance algorithm
portfolio approach. The related solver, {\sc satzilla}, can also
combine portfolio and multi-engine approaches. The algorithm portfolio approach is employed also
in: \cite{gome-selm-2001} on Constraint Satisfaction and
MIP,~\cite{samu-memi-2007} on QSAT and~\cite{gere-etal-2009} on
planning problems. If we consider
 ``pure'' approaches, the advantage of the algorithm portfolio over a
 multi-engine is that it is possible, by combining algorithms, to reach
 a performance than is better than the one of the best engine, 
 which is an upper bound for a multi-engine solver instead. On the other hand,
 multi-engine treats the engines as a black-box, and this is a fundamental
 assumption to have a flexible and modular system:
to add a new engine one just needs to update the inductive model.
Other approaches, an overview can be found in~\cite{hoos-2012}, 
work by designing methods for automatically tuning
and configuring the solver parameters: %
e.g.,~\cite{hutt-etal-2009,hutt-etal-2010} for solving SAT and MIP
problems, and~\cite{vall-etal-2011} for planning problems.

About the other approaches in ASP, the one implemented in
{\claspf}~\cite{gebs-etal-2011-claspfolio} mixes characteristics of
the algorithm portfolio approach with others more similar to this
second trend: it works by selecting the most promising {\clasp}
internal configuration on the basis of both ``static'' and ``dynamic''
features of the input program, the latter obtained by running {\clasp}
for a given amount of time. 
Thus, like the algorithm portfolio
approaches, it can compute both static and dynamic features, while trying to
automatically configure the ``best'' {\clasp} configuration on the
basis of the computed features. 

The work here presented is in a different ballpark w.r.t. 
\claspfolio for a number of motivations. First, from a machine
learning point of view, the inductive models of \measp are based on
{\it classification} algorithms, while the inductive models of
\claspfolio are mainly based on {\it regression} techniques, as
in {\sc satzilla}, with the
exception of a ``preliminary'' stage, in which a classifier is invoked
in order to predict the satisfiability result of the input instance.
Regression-based techniques usually need many training
instances to have a good prediction while, as shown in our paper,
this is not required for our method that is based on classification. 
To highlight consideration of the prediction power,
in Section~\ref{sub:discussion} we have applied our
approach to {\claspf}, showing that relying on classification instead of
regression in {\claspf} can lead to better results.
Second, as mentioned before, in our approach we consider the engines
as a black-box: \measp architecture is designed to be independent from
the engines internals. {\measp}, being a 
multi-engine solver, has thus higher modularity/flexibility w.r.t. 
{\claspfolio}: adding a new solver to {\measp} is immediate, while this is problematic
in {\claspfolio}, and likely would boil down to implement the
new strategy in {\clasp}. Third, as a consequence of the previous
point, we use only static features: dynamic features, as in the case
of \claspfolio, usually are both strongly related to a given engine and possibly costly to compute, 
and we avoided such kind of features. For instance, one of the \claspfolio dynamic
feature is related to the number of ``learnt constraints'', that could be a
significant feature for \clasp but not for other systems, e.g.,
\dlv that does not adopt learning and is based on look-ahead. 
Lastly, as described in Section~\ref{sub:feature}, we use only
cheap-to-compute features, while {\claspf} relies some quite
``costly'' features, e.g., number of SCCs and loops. This was confirmed on some
preliminary experiments: it turned out that {\claspf} feature
extractor could compute, in 600s, all its features for 573 out of 823
{\NP} ground instances.

An alternative approach in ASP is followed in the {\sc dors} framework
of Balduccini~\citeyear{bald-11}, where in the off-line learning
phase, carried out on representative programs from a given domain, a
heuristic ordering is selected to be then used in {\smodels} when
solving other programs from the same domain. The target of this work
seems to be real-world problem domains where instances have similar
structures, and heuristic ordering learned in some (possibly small)
instances in the domain can help to improve the performance on other
(possibly big) instances. According to its author\footnote{Personal
  communications with Marcello Balduccini.}  the solving method behind
{\dors} can be considered ``complementary'' more than alternative
w.r.t. the one of {\measp}, i.e., they could in principle be
combined. An idea can be the following: while computing features, one
can (in parallel) run one or more engines in order to learn a
(possibly partial) heuristic ordering. Then, in the solving phase,
engines can take advantage from the learned heuristic (but, of course,
assuming minimal changes in the engines). This would come up to having
two ``sources'' of knowledge: the ``most promising'' engine, learned
with the multi-engine approach, and the learned heuristic ordering.

Finally, we remark that this work is an extended and revised version of
\cite{mara-etal-2012-iclp}, the main improvements include: 
\begin{itemize}
\item[$(i)$] the adoption of six classification methods (instead of the only one, i.e.,
{\nn}, employed in \cite{mara-etal-2012-iclp}); 
\item[$(ii)$] a more detailed analysis of the dataset and the test sets;
\item[$(iii)$] a wider experimental analysis, including $(iiia)$ more systems, i.e., different
versions of {\measp} and {\claspf}, and $(iiib)$ more investigations on training and test sets, and 
\item[$(iv)$] an improved related work, in particular w.r.t.  the comparison with {\claspf}.
\end{itemize}

\section{Conclusion}\label{sec:conclusion}
In this paper we have applied machine learning techniques to ASP solving
with the goal of developing a fast and robust multi-engine ASP solver.
To this end, we have: 
$(i)$ specified a number of cheap-to-compute syntactic features 
that allow for accurate classification of ground ASP programs;
$(ii)$ applied six multinomial classification methods to
learning algorithm selection strategies; and
$(iii)$ implemented these techniques in our multi-engine solver \measp,
which is available for download at 
\begin{center}
\url{http://www.mat.unical.it/ricca/me-asp} .
\end{center}
The performance of \measp  was assessed on three experiments,
which were conceived for checking efficiency and robustness of our approach,
involving different training and test sets of instances taken 
from the ones submitted to the System Track of the 3rd ASP Competition.
Our analysis shows that our multi-engine solver {\measp} is very robust and efficient, 
and outperforms both its component engines and state-of-the-art solvers.

 \paragraph{Acknowledgments.} The authors would like to thank Marcello
 Balduccini for useful discussions (by email and in person) about the
 solving algorithm underlying his system {\dors}, and all the members
 of the {\claspf} team, in particular Torsten Schaub and Thomas Marius
 Schneider, for clarifications and the valuable support to train
 {\claspf} in the most proper way.


\newcommand{\SortNoOp}[1]{}

\end{document}